\documentclass[lettersize,journal]{IEEEtran}
\usepackage{amsmath,amsfonts}
\usepackage{algorithmic}
\usepackage{algorithm}
\usepackage{array}
\usepackage[caption=false,font=normalsize,labelfont=sf,textfont=sf]{subfig}
\usepackage{txfonts}
\usepackage{textcomp}
\usepackage{stfloats}
\usepackage{url}
\usepackage{verbatim}
\usepackage{graphicx}
\usepackage{cite}
\usepackage[square,sort&compress,comma,numbers]{natbib}
\usepackage[usenames,dvipsnames]{xcolor}
\usepackage[most]{tcolorbox}
\usepackage{textcomp}
\usepackage{multirow}
\usepackage{tabularx}
\usepackage{tikz}
\newcommand*\emptycirc[1][1ex]{\tikz\draw (0,0) circle (#1);} 
\newcommand*\halfcirc[1][1ex]{%
  \begin{tikzpicture}
  \draw[fill] (0,0)-- (90:#1) arc (90:270:#1) -- cycle ;
  \draw (0,0) circle (#1);
  \end{tikzpicture}}
\newcommand*\fullcirc[1][1ex]{\tikz\fill (0,0) circle (#1);} 

\newcolumntype{C}[1]{>{\centering\arraybackslash}p{#1}} 
\newcolumntype{L}[1]{>{\RaggedRight\arraybackslash}p{#1}} 
\newcolumntype{R}[1]{>{\RaggedLeft\arraybackslash}p{#1}} 
\usepackage{booktabs}
\usepackage{ragged2e} 
\usepackage{afterpage}
\usepackage{subcaption}
\usepackage{enumerate}
\usepackage{mathtools}
\usepackage{setspace}
\usepackage{hyperref}
\usepackage[figuresright]{rotating}
\usepackage{tablefootnote}

\newcommand{\paragraphb}[1]{\noindent{\bf #1} }

\begin{document}

\title{A Survey on Federated Unlearning: \\ Challenges and Opportunities}

\author{Hyejun Jeong, Shiqing Ma, Amir Houmansadr
}

\markboth{Journal of \LaTeX\ Class Files,~Vol.~14, No.~8, August~2021}%
{Shell \MakeLowercase{\textit{et al.}}: A Sample Article Using IEEEtran.cls for IEEE Journals}


\maketitle

\begin{abstract}
Federated learning (FL), introduced in 2017, facilitates collaborative learning between non-trusting parties with no need for participants to explicitly share raw data. This allows training a model on different private sets of user data while respecting privacy regulations such as GDPR and CPRA. However, emerging privacy requirements may mandate model owners to be able to \emph{forget} learned data, e.g., when requested by data owners or law enforcement. This has given birth to an active field of research called \emph{machine unlearning}. In the context of FL, many techniques developed for unlearning in centralized settings are not trivially applicable! This is due to the unique differences between centralized and distributed learning settings, specifically interactivity, stochasticity, heterogeneity, and limited accessibility in FL. In response, a recent line of work has focused on developing unlearning mechanisms tailored to FL. 

This survey paper aims to take a deep look at the \emph{federated unlearning} literature, with the goal of identifying research trends and challenges in this emerging field. By carefully categorizing papers published on FL unlearning (since 2020), we aim to pinpoint the unique complexities of federated unlearning, highlighting limitations on directly applying centralized unlearning methods. We compare existing methods regarding influence removal and performance recovery, compare their threat models and assumptions, and discuss their implications and limitations. We additionally analyze the experimental setup from various perspectives, including data heterogeneity and its simulation, datasets used for demonstration, and evaluation metrics. Our work aims to offer insights and suggestions for future research on federated unlearning.
\end{abstract}


\section{Introduction}
\IEEEPARstart{T}{he} ``Right to be Forgotten'' (RTBF) has gained attention upon its official recognition in 2014 \footnote{https://eur-lex.europa.eu/legal-content/EN/TXT/?uri=CELEX\%3A62012CJ0131}, aligning closely with privacy preservation mandates. 
The General Data Protection Regulation (GDPR) further stipulated the right to erasure \footnote{https://eur-lex.europa.eu/eli/reg/2016/679/oj}, empowering individuals to request personal data removal. 
Individuals may want their data to be forgotten from a trained model for various reasons, including privacy, security, or usability concerns. If they no longer consent to the privacy policy or terms of service, the data providers should be able to request the elimination of their explicit information as well as the influence on the model.
From the model owner's perspective, if they found a model is trained on compromised or faulty data, such as backdoor poisoned data samples, they would also want to remove this data to improve security and utility \cite{cao2015towards}. 
This has motivated an active line of work referred to as \emph{Machine Unlearning} (MU), aiming to remove (sensitive or faulty) data from trained ML models without retraining.

The trivial way to unlearn is to retrain the model from scratch, excluding the data to be removed. 
However, this is not only expensive in terms of overhead in time, memory, and resource consumption, but it may also not be feasible in various practical scenarios (e.g., the data to be unlearned are not open to the public).
Therefore, the community has developed unlearning mechanisms to efficiently erase the requested data and its influence on the model while minimally impacting the model's performance \cite{cao2023fedrecover}.
Unlearning steps begin during or after training or convergence and continue until evaluation metrics are satisfied, generating \emph{unlearned model} \cite{nguyen2022survey}.

\paragraphb{Unlearning in Federated Learning (FL):}
FL \cite{mcmahan2017communication} is a distributed machine learning framework that emerged in 2017 to comply with stringent privacy policies. 
It enables multiple clients to collaboratively train a model while maintaining their data locally. 
This paradigm preserves data privacy by design, reducing the risk of privacy breaches from eavesdropping or misuse by data-collecting entities.
In FL, a server initializes a global model and distributes the parameters to clients. 
Each client then trains a local model with private data and sends the trained model to the server, which aggregates the local models to update the global model. This iterative process continues until a stopping criterion is met.

In the context of FL, unlearning techniques developed for centralized settings are not trivially applicable! 
This is due to the unique differences between centralized and distributed learning, in particular, interactivity, stochasticity, heterogeneity, and limited accessibility. 
Once a model is trained in a federated manner, unlearning must also be performed in a federated setting. 
When a client requests to unlearn its data, it may perform centralized unlearning locally and send the updates to the server for aggregation. 
However, due to the interactivity inherent in FL, the unlearning effect can be trivial if the server simply averages all the local models. To address this, the target client, server, and remaining clients must collaborate to achieve proper unlearning, even with no access to the raw target data.
To this end, recent works have focused on unlearning mechanisms tailored to FL~\cite{liu2021federaser}.

\subsection{Contributions of This Survey}

This paper aims to provide a comprehensive view of the state of Federated Unlearning (FU) literature, offering insights and recommendations for future research on this topic. 
In summary, our contribution can be summarized as follows:

\begin{itemize}
    \item We highlight the unique complexities of unlearning in the federated setting, demonstrating the need for tailored unlearning mechanisms for FL (\autoref{sec:challenges}).
    \item We identify and compare the assumptions made in existing FU literature. This includes the entities undertaking unlearning, data distribution specifications, dataset usage, models utilized, and aggregation methods employed, and the research implications (\autoref{sec:assumptions}).
    \item We compare unlearning targets in different works (\autoref{sec:federatedunlearning}), categorize existing FU techniques regarding influence removal and performance recovery, and discuss their limitations (\autoref{sec:technique}). We also compare the evaluation objectives and metrics used (\autoref{sec:evalmetric}).
    \item We provide suggestions for future research on FU based on lessons and insights from our investigation of the existing FU literature (\autoref{sec:insights}).
\end{itemize}

\subsection{Comparison to Other Surveys in Unlearning}

\begin{table}
   \aboverulesep=0.2ex 
   \belowrulesep=0.4ex 
    \centering 
    \small
    \renewcommand*{\arraystretch}{0.7}
    \setlength\tabcolsep{4pt} 
    \caption{Comparisons to other surveys. Full, partial, and no coverage are represented by \fullcirc, \halfcirc, and \emptycirc, respectively.}
    \begin{tabular}{c|c|ccc|ccccccc|cc}
    
    \toprule
    \textbf{Ref.} & \textbf{Year} & \multicolumn{3}{c|}{\textbf{FU}} & \multicolumn{7}{c|}{\textbf{Unlearning}} & \multicolumn{2}{c}{\textbf{Insight}} \\ \midrule
    & & \begin{sideways}Federated context\end{sideways} 
    & \begin{sideways}Complexity specific to FL\end{sideways} 
    & \begin{sideways}Non-IID Data \& simulation\end{sideways} 
    & \begin{sideways}Taxonomy\end{sideways} 
    & \begin{sideways}Who unlearns\end{sideways} 
    & \begin{sideways}Implication\end{sideways} 
    & \begin{sideways}Influence removal\end{sideways} 
    & \begin{sideways}Performance recovery\end{sideways} 
    & \begin{sideways}Limitation\end{sideways} 
    & \begin{sideways}Evaluation metrics\end{sideways} 
    & \begin{sideways}Dataset statistics\end{sideways} 
    & \begin{sideways}Research direction\end{sideways} \\ \toprule 

    \cite{nguyen2022survey} & 2022 & \halfcirc & \halfcirc & \emptycirc & \fullcirc & \emptycirc & \fullcirc & \fullcirc & \emptycirc & \halfcirc & \fullcirc & \fullcirc & \fullcirc \\
    \cite{xu2023machine} & 2023 & \halfcirc & \halfcirc & \emptycirc & \fullcirc & \emptycirc & \emptycirc & \fullcirc & \emptycirc & \fullcirc & \fullcirc & \emptycirc & \fullcirc \\
    \cite{wu2023knowledge} & 2023 & \fullcirc & \fullcirc & \emptycirc & \emptycirc & \emptycirc & \emptycirc & \fullcirc & \emptycirc & \emptycirc & \fullcirc & \emptycirc & \fullcirc \\
    \cite{wang2023federated} & 2023 & \fullcirc & \emptycirc & \emptycirc & \emptycirc & \fullcirc & \emptycirc & \halfcirc & \emptycirc & \emptycirc & \emptycirc & \emptycirc & \emptycirc \\
    \cite{yang2023survey} & 2023 & \fullcirc & \fullcirc & \emptycirc & \fullcirc & \emptycirc & \fullcirc & \halfcirc & \emptycirc & \emptycirc & \fullcirc & \emptycirc & \fullcirc \\
    \cite{liu2023survey} & 2023 & \fullcirc & \fullcirc & \halfcirc & \fullcirc & \fullcirc & \emptycirc & \fullcirc & \halfcirc & \fullcirc & \fullcirc & \emptycirc & \fullcirc \\
    \cite{shaik2024exploring} & 2024 & \halfcirc & \emptycirc & \emptycirc & \fullcirc & \emptycirc & \emptycirc & \fullcirc & \emptycirc & \emptycirc & \fullcirc & \fullcirc & \fullcirc \\
    \cite{romandini2024federated} & 2024 & \fullcirc & \fullcirc & \fullcirc & \fullcirc & \fullcirc & \emptycirc & \fullcirc & \emptycirc & \fullcirc & \fullcirc & \emptycirc & \fullcirc \\
    \textbf{Ours} & 2025 & \fullcirc & \fullcirc & \fullcirc & \fullcirc & \fullcirc & \fullcirc & \fullcirc & \fullcirc & \fullcirc & \fullcirc & \fullcirc & \fullcirc \\
    \bottomrule
    \end{tabular}
    \label{tab:comparisontotheothers}
    \vspace{-10pt}
\end{table}

In \autoref{tab:comparisontotheothers} we compare several recent surveys on Machine Unlearning \cite{nguyen2022survey, xu2023machine, shaik2024exploring} and Federated Unlearning \cite{wu2023knowledge, wang2023federated, liu2023survey, yang2023survey} regarding various dimensions. 
As can be seen, while some surveys have considered unlearning in the federated context, not all addressed the unique complexities of FU, often overlooking factors such as data heterogeneity in FL. Furthermore, because FU involves multiple parties with varying knowledge and capabilities, disclosing raw original and unlearning model parameters to all participants could potentially facilitate adversaries in inferring the removed information. Thus, identifying the parties involved in unlearning and implementing robust safeguards is one of the necessary aspects to be considered, although only a few surveys \cite{wang2023federated, liu2023survey, romandini2024federated} discussed it.

\begin{figure}[t]
\centerline{\includegraphics[width=\linewidth]{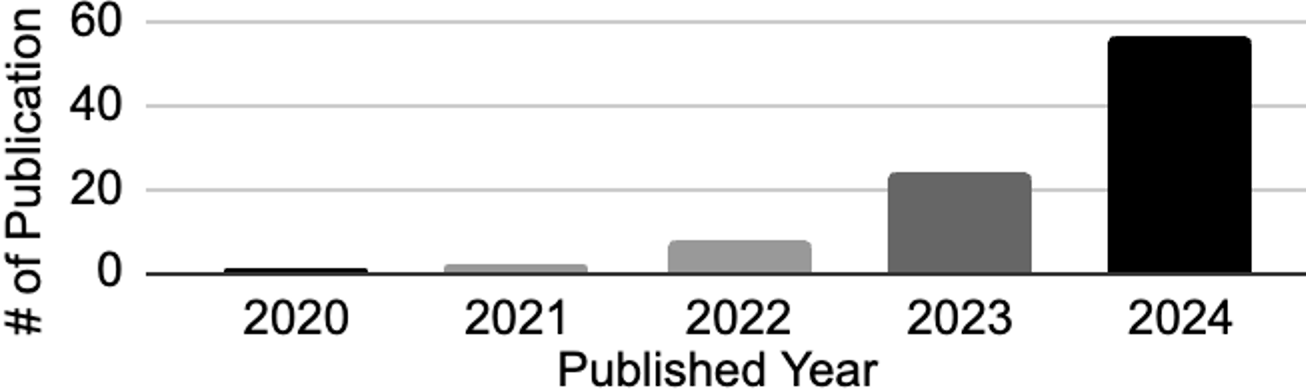}}
\caption{Number of Federated Unlearning Publications.}\label{fig:numpublication}
\vspace{-5pt}
\end{figure}

Compared to the standard machine unlearning, federated unlearning is in its infancy. 
By searching on Google Scholar, we try to identify works related to the keywords ``unlearning'' and ``federated'' or ``distributed'' at the same time. \autoref{fig:numpublication} summarizes the number of FU papers published each year until 2024 (i.e., 85 by 2024, 94 by March 2025). We only include, for example, LLM unlearning literature, only demonstrated on the federated context, as this work focuses on federated unlearning.
Although not exhaustive, the table reflects the exponentially increasing attention to FU, motivating our survey on identifying research gaps and challenges for future research.

\subsubsection{Comparison to a concurrent survey \cite{liu2023survey}}
The work most closely aligned with ours is \cite{liu2023survey}, primarily reviewing existing FU methods and evaluation metrics.
As our survey aims ultimately to provide insights for future work, we identify room for improvement in addition to analyzing existing literature in diverse dimensions. 
Specifically, our work delves deep into the practicality of each literature, including assumptions on data distribution and its simulation, along with dataset usage and research implications. 
We extract key takeaways from each part while exploring various concepts, properties, and methodologies.
At the end of the paper, we provide valuable insights and potential directions based on the takeaways for readers' clear understanding.

\section{Preliminaries}
\label{sec:preliminary}
We provide preliminaries on FU, including definitions, brief introductions to FL and MU, and the inherent complexities.

\subsection{Federated Learning}

\begin{figure}
\centerline{\includegraphics[width=\linewidth]{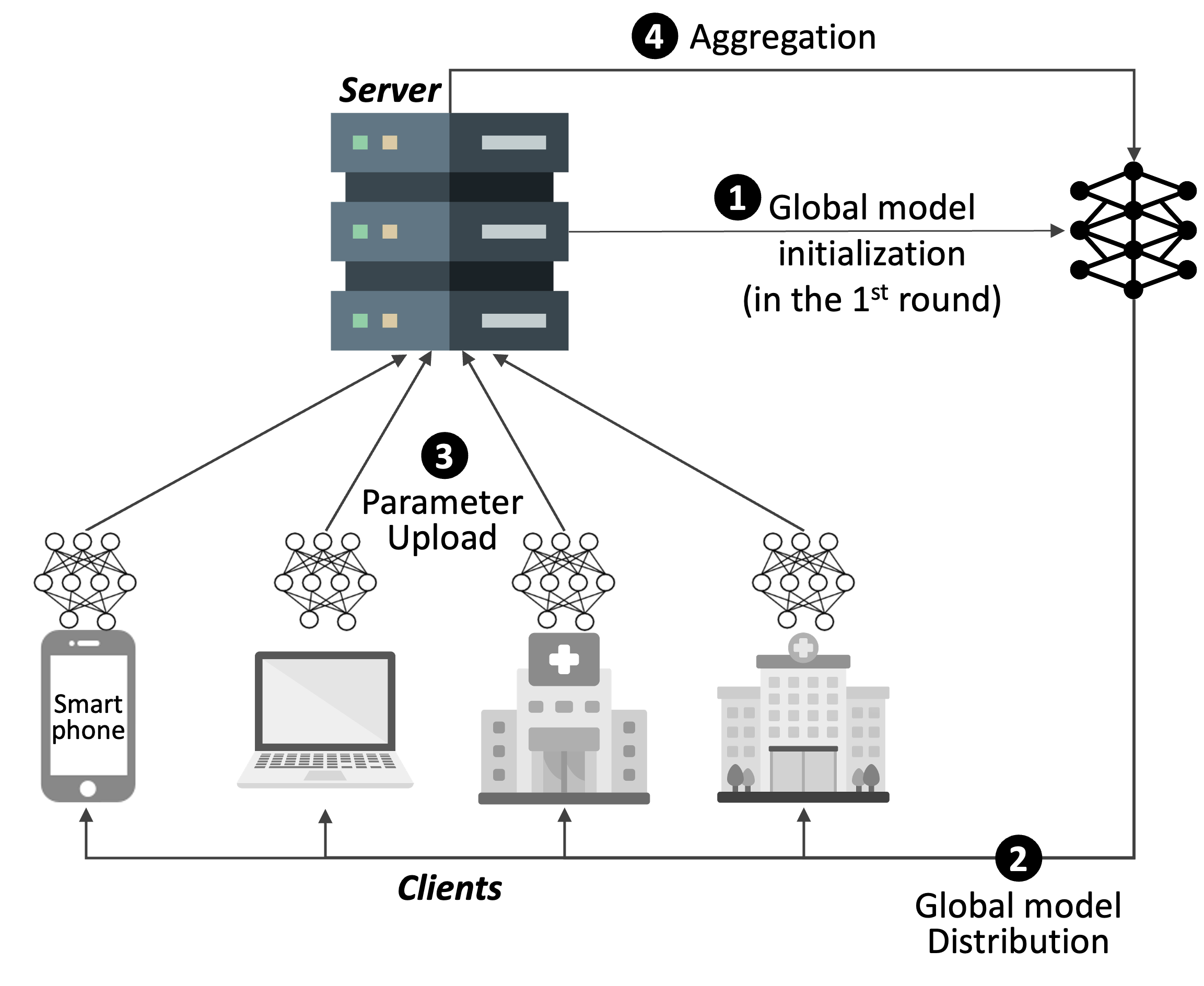}}
\caption{FL Training Workflow.}
\label{fig:flworkflow}
\vspace{-10pt}
\end{figure}

As illustrated in \autoref{fig:flworkflow}, FL \cite{mcmahan2017communication} trains a model in a privacy-preserving way by distributing model updates to the participating clients instead of collecting sensitive data in a centralized location. 
The basic FL workflow is as follows:

\begin{enumerate}
    \item [\textcircled{\small{1}}] A central server initializes a global model $w^0$.
    \item [\textcircled{\small{2}}] The server advertises the global model $w^t$ and selects $K=C\times N$ clients, a random fraction $C$ of clients out of a total of $N$ clients that will participate in the current training round. 
    \item [\textcircled{\small{3}}] Each $K$ client computes local updates $g_k^t$ (as Eq.~\ref{eq:updates}) with its training dataset $\mathcal{D}_k$ based on the current global model $w^t$. 
    Then, they send the updates $g_k^t$ to the server. 
        \setlength{\abovedisplayskip}{1pt}%
        \setlength{\belowdisplayskip}{1pt}%
        \setlength{\abovedisplayshortskip}{1pt}%
        \setlength{\belowdisplayshortskip}{1pt}%
        \setlength{\jot}{1pt}
        \begin{align} 
             F_k(w^t) &= \frac{1}{D_k} \sum_{d \in \mathcal{D}_k} f_d(w_k^t) \\
             g_k &= \nabla F_k(w_t)
            \label{eq:updates}
        \end{align}
        where $\mathcal{D}_k$ is the set of data points with a size of $D_k$ a client $k$ has, $w_k^t$ is the local model weight at round $t$, $f_d$ is the loss of the prediction on sample $d$, and $F_k$ is the average loss over the set of data points.
        
    \item [\textcircled{\small{4}}] The server aggregates $K$ received updates and creates an updated global model $w^{t+1}$ (as Eq. \ref{eq:updated_global}).
    \begin{equation}
     w^{t+1} \leftarrow w^t - \eta \frac{1}{K}\sum_{k=1}^K g_k
    \label{eq:updated_global}
    \end{equation}
    where $\eta$ is the learning rate.
\end{enumerate}


\textcircled{\small{2}} to \textcircled{\small{4}} are repeated until the pre-defined stopping criterion is met. 
FL algorithms are named differently based on how they collect and process updates to create the global model.
\texttt{FedAvg} \cite{mcmahan2017communication}, for instance, is a foundational FL algorithm that averages the local model parameters. Whereas \texttt{FedBuff} \cite{nguyen2022federated} is a more recent method designed to reduce communication overhead and handle heterogeneous data by buffering updates.

\paragraphb{Components.} A global server and multiple local clients comprise an FL process. The server, assumed to be equipped with more powerful computation and memory resources, is responsible for aggregating local clients' model parameters to update the global model. Nonetheless, as it cannot access raw data, its knowledge is limited to the model parameters but not training data. Similarly, local clients can access their own training dataset, not each other's. Key terminologies related to FL are summarized in \autoref{tab:terminology} in \autoref{app:fl_terminology}.

\subsection{Machine Unlearning} 
MU aims to remove specific samples, classes, or features from a trained model upon request. A naive approach is to retrain the model from scratch, excluding the data to forget \cite{cao2015towards}. Nonetheless, this is often impractical due to its overhead, especially when the model has been trained with a huge dataset. As such, efficient unlearning methods, faster than \textit{retraining}, have been explored, effectively forgetting the data while maintaining performance on remaining data. The objective is to make the unlearned model perform as similarly as possible to the retrained one. 
MU methods are broadly categorized into data-driven \cite{nguyen2022survey, xu2023machine} approaches including data partition \cite{bourtoule2021machine}, obfuscation \cite{graves2021amnesiac, cao2015towards}, or augmentation \cite{huang2021unlearnable}, and model manipulation \cite{peste2021ssse, xu2023machine} approaches include model shifting, pruning, and replacement. 

The evaluation metrics are various \cite{nguyen2022survey}. Broadly categorizing unlearning effectiveness metrics, there are performance-based (accuracy or error rates), distance-based (KL divergence, L2 distance, or cosine similarity), and attack-based (attack success rate of backdoor or membership inference attacks) metrics. Many of these metrics overlap with those used for FU. Detailed evaluation metrics are explained in later sections. 

\paragraphb{Complexities in Machine Unlearning.} \label{subsubsec:mu_complexities}
The primary challenge in MU is to maintain the model's performance on remaining data while completely forgetting specific knowledge. 
Erasing the sample (\textit{target removal}) is straightforward, but eliminating its influence is uneasy. The stochastic nature of training with randomly ordered batches, incremental nature of training, and catastrophic forgetting leads to significant performance degradation on the remaining data \cite{nguyen2022survey}.

\section{Introducing Federated Unlearning}
\label{sec:federatedunlearning}

\begin{figure*}[tp]
\centerline{\includegraphics[width=\linewidth]{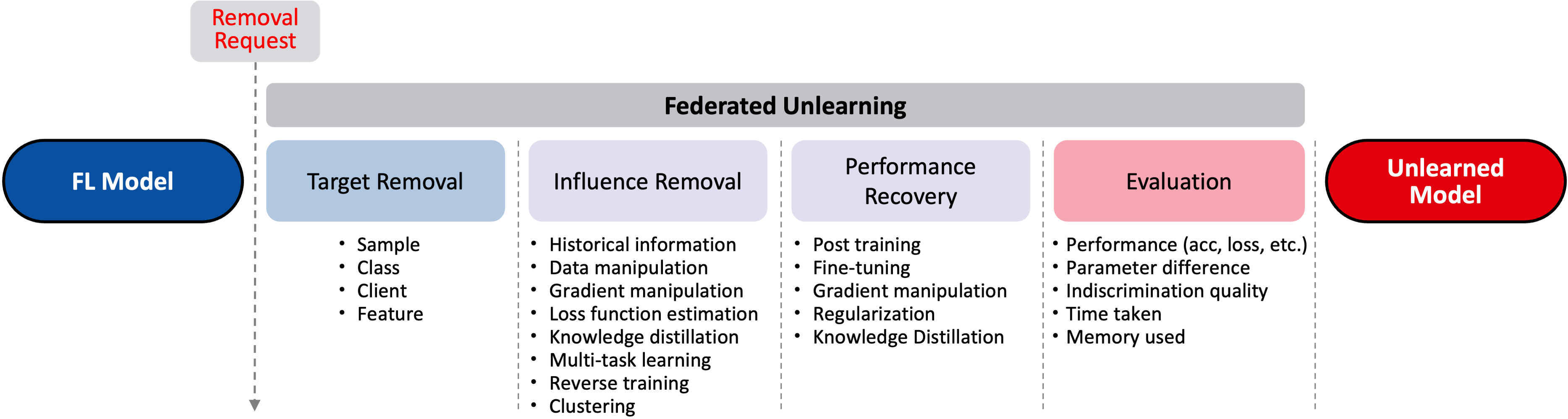}}
\caption{Federated Unlearning Workflow. The server or clients initiate target removal during or after the FL process. The unlearner excludes the target, erases its contribution, and recovers performance. The requesting client verifies the proper elimination using evaluation metrics, generating an unlearned model.}
\label{fig:overview}
\vspace{-5pt}
\end{figure*}

\autoref{fig:overview} depicts the overall FU flow. During or after training an FL model, either a server or clients request to remove certain target information. 
If the server initiates unlearning, for instance, to remove low-quality data, it additionally takes a step to identify the low-performing client to be deleted. Otherwise, the target samples or clients are removed from the learning process upon request. 

\paragraphb{What is Unlearned?}
Unlearning target refers to the data or information we want the model to forget. Upon request, the target(s) are removed from the training set (\textit{target removal}). Although some methods \cite{gu2024unlearning, wu2022federated, meerza2024confuse, zuo2024federated, liu2024blockful, varshney2024efficient, wang2025forgettingdatatimetheoretically, Zhong2025UnlearningTK, gong2022compressed} unlearn knowledge in multiple granularity, all client, class, and samples or two of them, most of the works focus on unlearning (a) client(s) because of the introduction of it in the federated context, unlike the centralized setting. 
We explain the different granularities of knowledge that are unlearned. 

\textit{\textbf{Client}} removal involves erasing all data owned by specific clients from the model \cite{liu2022right, pan2022machine, liu2021federaser, zhang2023fedrecovery, su2023asynchronous, xu2023revocation, jiang2024efficient, varshney2024efficient, chundawat2024conda, wu2024unlearning, sheng2024robust, liu2024guaranteeing, huynh2024fast, su2024f2ul, chen2024upcycling, deng2024enable, pan2024federated, wang2024fedu, li2024federated, ma2024hier, yu2024federated, wang2024efficient, zhu2024federated, xie2024adaptive, mora2024fedunran, zuo2024federated, jiang2024feduhb, mora2024fedquit, xiong2024appro,khalil2025not, ameen2025speed, zhang2025model, wang2025poisoning, huynh2025certified}. It overlaps with sample removal if all data owned by a specific client are subject to be removed \cite{cao2023fedrecover, gao2024verifi, yuan2023federated, jin2023forgettable, li2023federated, zhao2023federated, guo2023fast, wang2023mitigating}, and class removal when each client has a distinct class of samples \cite{gong2022forget, ye2023heterogeneous}. 

\textit{\textbf{Class}} removal aims at removing all samples labeled as the target class. Think of label flipping or backdoor attacks; compromised clients hold trigger-injected samples labeled as a certain class (e.g., ``9'') while other clients hold normal data. Here, the target to unlearn would be the class 9 \cite{ameen2024addressing, sheng2024robust, xu2024update, guo2024forgetting}. Depending on sample distribution across clients, it can also be treated as a client removal problem \cite{gong2022compressed, dhasade2024quickdrop}. 

\textit{\textbf{Sample}} removal involves eliminating particular data instances (e.g., removing mistakenly uploaded or private photos and their influence) \cite{xia2023fedme, elbedoui2023ecg, zhou2024streamlined}. It includes sanitizing the training dataset by removing polluted samples, such as compromised \cite{wang2023bfu, alam2023get, li2023subspace, li2023federated, wu2024unlearning, ameen2024addressing, wang2024server} or outdated samples \cite{xu2023machine, wu2023knowledge, wang2023edge}. 

\textit{\textbf{Feature}} removal is forgetting specific sensitivity in training data \cite{gu2024ferrari}. It can be used to improve model generalizability; for example, features could represent data heterogeneity, as seen in \cite{dinsdale2022fedharmony}, that removed the scanner bias from medical images collected from different sites. It also overlaps with sample removal when eliminating samples with poison-indicating features and is analogous to client removal in vertical or cross-silo FL that each client has distinct features \cite{liu2021revfrf, deng2023vertical, zhang2023securecut, han2025vertical, pan2025feature}.

\section{Unique Challenges of Federated Unlearning}
\label{sec:challenges}

The federated context brings unique complexities to FU, in addition to those in centralized unlearning, described in \autoref{subsubsec:mu_complexities}. In this section, we explore why MU methods cannot be directly applied to FU and investigate the inherent difficulties involved. 
Note that the complexities are not orthogonal nor mutually exclusive but usually related; to address one concern, the other might also need to be resolved. It makes complete FU much more complex to achieve. 

\textbf{Interactive training} embeds information in a model \cite{liu2021federaser, wu2023knowledge}. The subsequent training rounds accumulate all the information propagated throughout global and local models \cite{liu2022right}. In addition to the information forwarding coupling in parameter updates, interactivity further adds many more difficulties. FL interactively trains a model by iteratively aggregating local models on the server, diminishing the impact of a local model upon aggregation. It then means that removing the information and influence from only one local model, analogous to that in MU, is no longer feasible in the federated context as it soon becomes ineffective after it is aggregated. 

\textbf{Information isolation} characterizes FL that allows privacy-preserving training by keeping data local to clients, unlike centralized (un)learning, where a single entity has access to all data. FU involves three parties with varying levels of data accessibility: the server, target clients, and remaining clients. When the server or remaining clients perform unlearning, it is different from centralized unlearning and more complicated as they do not have access to raw data. Moreover, unlearning methods utilizing raw data (e.g., data partition \cite{bourtoule2021machine}, augmentation \cite{huang2021unlearnable}, or influence \cite{peste2021ssse}) become inapplicable due to limited access to local data from the server or other clients \cite{nguyen2022survey, wu2023knowledge}.

\textbf{Non-IID data} across clients, one of the famous open problems in FL \cite{kairouz2021advances}, adds difficulty to FU. In contrast to centralized ML, each FL client separately trains a local model based on its training data that could vary in distribution.
This heterogeneity results in some clients holding unique or skewed data distributions, causing their updates to disproportionately impact specific model parameters. It complicates isolating and eliminating the influence of target data, as parameter updates reflect a blended contribution from multiple clients.

\textbf{Stochastic client selection} introduces an additional layer of uncertainty, as the server receives updates from arbitrarily selected clients in each training round \cite{liu2022right, wu2023knowledge}. This extends the stochasticity seen in MU with randomly ordered batch \cite{nguyen2022survey}. Since a client's impact is spread across non-contiguous rounds, pinpointing when and how to modify model parameters becomes challenging. Moreover, on-and-off client participation causes delayed propagation of updates, meaning the unlearned impact can persist in subsequent rounds, even when they are not actively participating. While frequent checkpointing or rollback can address this, it is highly resource-intensive, especially if the client participated early in training.

\begin{table}[htbp]
    \centering 
    \setlength\tabcolsep{2.8pt} 
    \caption{Accessible knowledge depending on who unlearns. ``Global,'' ``Own local,'' ``All local,'' and ``Raw data'' refer to the global model updates, own local model updates, all local model updates, and raw data to forget, respectively.}
    \begin{tabular}{ccccc}
    \toprule
        \textbf{Unlearner} &\textbf{Global} & \textbf{Own local} & \textbf{All local} & \textbf{Raw data} \\ \toprule
        Server              & \checkmark &            & \checkmark & \\
        Target client       & \checkmark & \checkmark &            & \checkmark \\
        Remaining clients   & \checkmark & \checkmark &            & \\
        \bottomrule
    \end{tabular}
    \label{tab:accessibility}
    \vspace{-10pt}
\end{table}

\begin{tcolorbox}[enhanced, breakable, fontupper=\linespread{.93}\selectfont]
    \textbf{\hypertarget{t1}{T-1.} Takeaways on Unique Challenges of FU.}
    Unlike MU, which unlearns information within accessible raw data in a central location, FU cannot solely rely on the target client due to the diminishing impact of individual clients upon aggregation. Isolated information, limited accessibility to data and its distribution, and extra uncertainty stemming from client selection introduce additional challenges to FU, as not all participants consistently have access to raw data.
\end{tcolorbox}

\section{Comparison of Existing System Models in FU}
\label{sec:assumptions}

We overview and compare the system models used in the existing FU literature from various dimensions. 

\subsection{Who Unlearns?}

``Who unlearns'' differentiates FU from MU and FL. In the federated context, the entities performing unlearning have knowledge and capabilities that depend on who they are. ~\autoref{tab:accessibility} summarizes the knowledge available to each unlearner. Each client has its local and global model updates but no updates from the others. Thus, if only the remaining clients unlearn, they must rely on the global model updates they have received. When the target client performs unlearning, it can additionally leverage its raw training data. On the other hand, if the server performs unlearning, it has access to the global model and updates from all clients' models. As indicated in the ``Unlearner'' column of ~\autoref{tab:whounlearns}, most works chose to perform unlearning on the target clients, given their access to raw data to unlearn. The server, accessible to all local updates and identifying target clients, often takes charge of unlearning. Nonetheless, in scenarios where target clients request removal and leave, remaining clients often cooperate with the server.

\begin{tcolorbox}[enhanced, breakable, fontupper=\linespread{.93}\selectfont]
    \textbf{\hypertarget{t2}{T-2.} Takeaways on Who Unlearns.}
    The clients' flexibility to dynamically join and leave poses a risk of malicious clients joining FU. Furthermore, if the target client has left the system and/or the unlearning performer is not the target client, unlearning would not be properly completed, leaving residuals to the early-leaving client. Consequently, it could leave room for potential exploitation by compromised parties, who unlearn in what way matters to assure complete removal, distinct from FL. 
\end{tcolorbox}

\begin{table}
    \centering 
    \footnotesize
    \setlength\tabcolsep{3pt} 
    \renewcommand*{\arraystretch}{0.8}
   \aboverulesep=0.2ex 
   \belowrulesep=0.4ex 
    \caption{Unlearner. Svg, Tgt, and Rmn refer to the server, target, and remaining clients, respectively.}
    \begin{tabular}{rccc|rccc}
    \toprule
    \multicolumn{1}{c}{\multirow{2}*{\textbf{Ref.}}} & \multicolumn{3}{c|}{\textbf{Unlearner}} & \multicolumn{1}{c}{\multirow{2}*{\textbf{Ref.}}} & \multicolumn{3}{c}{\textbf{Unlearner}} \\ [-.5ex]\cmidrule{2-4}\cmidrule{6-8}
    & Svr & Tgt & Rmn & & Svr & Tgt & Rmn \\ \toprule
    RevFRF\cite{liu2021revfrf} & $\medbullet$ & & & 2F2L\cite{jin2023forgettable} & & $\medbullet$ \\	
    Exact-Fun\cite{xiong2023exact} & & $\medbullet$ & $\medbullet$ & Liu et al.\cite{liu2022right} & & $\medbullet$ \\
    FedUHB\cite{jiang2024feduhb} & & & $\medbullet$ &    Xie et al.\cite{xie2024adaptive} & $\medbullet$ \\
    FATS\cite{tao2024communication}& $\medbullet$ & & &  Li et al.\cite{li2024federated} & $\medbullet$ & $\medbullet$ \\
    R2S\cite{wang2024efficient} & $\medbullet$ & & &    BadUnlearn\cite{wang2025poisoning} & $\medbullet$\\
    Zuo et al.\cite{zuo2024federated} & & & $\medbullet$ & Appro-Fun\cite{xiong2024appro} & & $\medbullet$ \\
    FedADP\cite{jiang2024efficient} & $\medbullet$ & & &  FedU\cite{wang2024fedu} & & $\medbullet$ \\
    Shao et al.\cite{shao2024federated} & $\medbullet$ & & &  FedME2\cite{xia2023fedme} & & $\medbullet$ & $\medbullet$ \\
    Wang et al.\cite{wang2024forget}  & & & $\medbullet$ & Alam et al.\cite{alam2023get} & & $\medbullet$ \\
    FedRecover\cite{cao2023fedrecover} & $\medbullet$ & & $\medbullet$ & BFU\cite{wang2023bfu} & & $\medbullet$ \\
    Wu et al.\cite{wu2024unlearning} & $\medbullet$ & & &  FedHarmony\cite{dinsdale2022fedharmony} & & $\medbullet$ & $\medbullet$ \\	
    Fu et al.\cite{fu2024client} & $\medbullet$ & & &  Chen et al.\cite{chen2024federated} & $\medbullet$ \\
    FedRecovery\cite{zhang2023fedrecovery} & $\medbullet$ & & &  Goldfish\cite{wang2024goldfish} & $\medbullet$ \\	
    CFRU\cite{huynh2025certified} & $\medbullet$& & &  FedLU\cite{zhu2023heterogeneous} & $\medbullet$ & $\medbullet$ & $\medbullet$ \\	
    FedADP\cite{wang2024efficient} & $\medbullet$ & & &  FedAF\cite{li2023federated} & & $\medbullet$ \\
    MetaFul\cite{wang2023mitigating} & $\medbullet$ & & &  FedQUIT\cite{mora2024fedquit} &  & $\medbullet$ \\
    Deng et al.\cite{deng2023vertical} & $\medbullet$ & & &      CKGD \cite{zhang2025model} & $\medbullet$ \\
    Starfish\cite{liu2024privacy} & $\medbullet$$\medbullet$\footnotemark & & & SFU\cite{zhou2024streamlined} & $\medbullet$ \\
    Crab\cite{jiang2024towards}  & $\medbullet$ & & &      HDUS\cite{ye2023heterogeneous} & & & $\medbullet$ \\
    FedEraser\cite{liu2021federaser} & & & $\medbullet$ &    FCU\cite{deng2024enable} & & $\medbullet$ \\
    FRU\cite{yuan2023federated} & $\medbullet$ & $\medbullet$ & $\medbullet$ &    Ferrari\cite{gu2024ferrari} & $\medbullet$ & $\medbullet$ \\
    SIFU\cite{fraboni2024sifu} & $\medbullet$ & & $\medbullet$ &    Imba-ULRc\cite{yu2024federated} & $\medbullet$ & & $\medbullet$\\
    FedRemover\cite{yuan2024towards} & $\medbullet$  & & &      Xu et al.\cite{xu2024update} & & & $\medbullet$\\
    FedUnlearn\cite{nguyen2024empirical} & $\medbullet$ & & &      TrustChain\cite{zuo2024federated} & $\medbullet$ \\
    FedUNRAN\cite{mora2024fedunran} & & $\medbullet$ &&  	    FUSED\cite{Zhong2025UnlearningTK} & $\medbullet$ & & $\medbullet$\\
    FUCRT\cite{guo2024forgetting} & & $\medbullet$ & $\medbullet$ &    EWC-SGA\cite{wu2022federated}& & $\medbullet$ \\
    FedMUA\cite{chen2025fedmua} & & $\medbullet$ &&    SFU\cite{li2023subspace} & $\medbullet$ & $\medbullet$  & $\medbullet$ \\	
    Pan et al.\cite{pan2025feature} &  $\medbullet$	& & &      Halimi et al.\cite{halimi2022federated} & & $\medbullet$ \\
    SecForget\cite{liu2020learn} & & $\medbullet$ &&	    QuickDrop\cite{dhasade2024quickdrop} & & $\medbullet$  & $\medbullet$ \\
    FFMU\cite{che2023fast} & & $\medbullet$ && LMR\cite{ameen2024addressing} & $\medbullet$ & $\medbullet$ \\
    FedFilter\cite{wang2023edge} & $\medbullet$ & & &  	    Wang et al.\cite{wang2024server} & & $\medbullet$ \\
    UKRL\cite{xu2023revocation} & & $\medbullet$ &&    Han et al.\cite{han2025vertical} & & $\medbullet$ \\
    MoDe\cite{zhao2023federated} & $\medbullet$ & $\medbullet$ &&    BlockFUL\cite{liu2024blockful} & & $\medbullet$ \\
    ConDa\cite{chundawat2024conda} & $\medbullet$ & & &      FedOSD\cite{pan2024federated} & $\medbullet$ \\
    RobustFU\cite{sheng2024robust} & $\medbullet$ & $\medbullet$ & $\medbullet$  &	    F2ul\cite{su2024f2ul} & $\medbullet$ \\
    FRAMU\cite{shaik2024framu} & & $\medbullet$ & $\medbullet$ &    VFU\cite{varshney2025unlearningclientsfeaturessamples} & & $\medbullet$ & $\medbullet$ \\
    VeriFi\cite{gao2024verifi} & $\medbullet$ & & &  	    forgetSVGD\cite{gong2022forget} & $\medbullet$ & & $\medbullet$ \\
    Lin et al.\cite{lin2024blockchain} & $\medbullet$ & & &  	    CforgetSVGD\cite{gong2022compressed} & & $\medbullet$\\
    FC\cite{pan2022machine} & $\medbullet$ & & &      KNOT\cite{su2023asynchronous} & $\medbullet$ & $\medbullet$ & $\medbullet$ \\
    Wang et al.\cite{wang2022federated} & $\medbullet$ & $\medbullet$ & $\medbullet$ &	    Lin et al.\cite{lin2024scalable} & & & $\medbullet$ \\
    SecureCut\cite{zhang2023securecut} & & $\medbullet$ &&    Lin et al.\cite{lin2024incentive} & $\medbullet$ \\
    FAST\cite{guo2023fast} & $\medbullet$ & & &  	    Liu et al.\cite{liu2024guaranteeing} & & & $\medbullet$ \\
    ElBedoui et al.\cite{elbedoui2023ecg} & & $\medbullet$ &&	    Hier-FUN\cite{ma2024hier} & & $\medbullet$\\ 
    Fast-FedUL\cite{huynh2024fast} & $\medbullet$ & & &      FedUMP\cite{zhu2024federated} & $\medbullet$ \\
    NoT\cite{khalil2025not} & $\medbullet$ & & &      k-IPfedAvg\cite{varshney2024efficient} & $\medbullet$  \\
    Wang et al.\cite{wang2025forgettingdatatimetheoretically} & $\medbullet$ & & &      FedAU\cite{gu2024unlearning} & $\medbullet$ & $\medbullet$ \\
    \bottomrule 
    \multicolumn{8}{l}{\footnotesize \textsuperscript{4} Uses two servers for two-party computation, for security}  
    \end{tabular}
    \label{tab:whounlearns}
    \vspace{-10pt}
\end{table}

\subsection{What Data Distribution?}

Another crucial consideration is ``under what data distribution,'' which is how data are distributed across clients. Data distribution and its simulation are summarized in the column ``Data Dist.'' and ``NIID sim.'' of ~\autoref{tab:settings}. Heterogeneous data, a more practical assumption in the distributed nature, often interferes with FL models from convergence \cite{zhao2018federated}. Similarly, an FU method designed for and demonstrated with IID data only would be highly unlikely to work as expected in practice with real-world data. Some work \cite{pan2022machine, su2023asynchronous, wang2023edge} naturally achieved it using real-world datasets, such as FEMNIST, Tiny-Shakespeare, or MovieLens-100k. The majority of the authors adopted prior probability shifts, as well as covariate shifts and concept drift. Below is a brief introduction to the three types and their implementation. For a more detailed description of non-IID data, refer to Section 3.1 in \cite{kairouz2021advances}. 

\textbf{Prior probability shift}, a variation in label distributions across clients, is often simulated using three primary methods:
\begin{itemize}
    \item \textbf{Dirichlet Distribution:} This method quantifies the degree of non-IIDness using a concentration parameter, $\alpha$ (within the interval $(0, \infty)$\footnote{For a visual representation, refer to Figure 4 in \cite{tang2023fedrad}}. As $\alpha \rightarrow \infty$, it mimics IID, while $\alpha \rightarrow 0$ imitates a higher degree of non-IID. Widely recognized for its ability to emulate real-world scenarios, many works employed this approach \cite{li2023subspace, su2023asynchronous, zhao2023federated, fraboni2024sifu, gao2024verifi, dhasade2024quickdrop, wang2023mitigating, gu2024ferrari, deng2024enable, wu2024unlearning, gu2024unlearning, guo2024forgetting, mora2024fedquit, khalil2025not, zhang2025model, wang2025poisoning, wang2025unlearning, chen2025fedmua}.
        
    \item \textbf{Fang's Method \cite{fang2020local}:} In this unique approach, clients are divided into $L$ groups, where $L$ denotes the number of classes in a dataset. Training instances with label $l$ are assigned to the $l$-th group with probability $p$ and to any other group with a probability of $\frac{1-p}{L-1}$. This method is adopted by \cite{cao2023fedrecover, wang2022federated}.
    
    \item \textbf{Unique Class Assignment:} \cite{gong2022forget, gong2022compressed, ye2023heterogeneous, pan2024federated} assign unique classes to each client such that each client exclusively holds data samples of one or two classes out of ten.
\end{itemize}
    
\textbf{Covariate shift}, or feature distribution skew, occurs when each client has different features of data instances with the same label shared across clients. An example is the dataset that FedHarmony \cite{dinsdale2022fedharmony} used. They wanted to unlearn site-specific bias from medical images so that the model could focus solely on biological features.

\textbf{Concept drift} refers to the changing input data distribution over time. When a sample is removed from the training dataset, the distribution shifts from its original state. With each subsequent sample removal, the distribution changes again, transitioning to a different state. FRAMU \cite{shaik2024framu} views unlearning as a continual learning process from distinct training datasets after each removal, so it adapts to these dynamic and evolving distributions.

\begin{tcolorbox}[enhanced, breakable, fontupper=\linespread{.93}\selectfont]
\textbf{\hypertarget{t3}{T-3.} Takeaways on Data Distribution.}
    More works assumed heterogeneous data in the federated context, although some undisclosed their specific simulation methods. The primary choice is employing Dirichlet distribution because of its recognized ability to mimic real-world datasets. However, as what value reflects the real world the most remains unclear, $\alpha$, controlling degree of non-IID, varies by a large margin across literature, limiting fair comparison across works. 
\end{tcolorbox}

\begin{table*}[htbp]
    \centering 
    \footnotesize
    \renewcommand*{\arraystretch}{0.76}
    \setlength\tabcolsep{5.4pt} 
    \captionsetup{skip=2pt}
    \caption{Data Dist. (distribution), NIID sim. (Non-IID simulation), data types, aggregation method, and implications. ``Dirichlet,'' ``Fang,'' ``unique,'' and ``random'' refer to Dirichlet distribution, Fang \cite{fang2020local}'s approach, limited class assignments, and random assignment; ``n/d'' and ``-'' denotes not disclosed and none;. \textbf{img, tab, txt}, and \textbf{oth} refer to the image, tabular, text, and other datasets. \textbf{efc, fid, efn, sec, gua, ada}, and \textbf{sca} represent efficacy, fidelity, efficiency, security, (theoretical) guarantee, adaptivity, and scalability. `Med' and `TrMean' indicate Median, Trimmed mean\cite{yin2018byzantine}, respectively.}
    \label{tab:settings}
    \begin{tabular}{rrrcccclccccccc}
    \toprule
    \multicolumn{1}{c}{\multirow{2}*{\textbf{Ref.}}} & \multicolumn{1}{c}{\multirow{2}*{\textbf{Data Dist.}}} & \multicolumn{1}{c}{\multirow{2}*{\textbf{NIID Sim.}}} & \multicolumn{4}{c}{\textbf{Data Type}}  & \multicolumn{1}{c}{\multirow{2}*{\textbf{Aggregation Method}}} & \multicolumn{7}{c}{\textbf{Implication}} \\ [-.5ex] \cmidrule{4-7}\cmidrule{9-15}
    & & & \textbf{img} & \textbf{tab} & \textbf{txt} & \textbf{oth} & & 
     \textbf{efc} & \textbf{fid} & \textbf{efn} & \textbf{sec} & \textbf{gua} & \textbf{ada} & \textbf{sca} \\ \toprule

    Exact-Fun \cite{xiong2023exact} & Non-IID & Random & $\medbullet$ & & &  & FedAvg & $\medbullet$ & $\medbullet$ & $\medbullet$ \\
    FedUHB \cite{jiang2024feduhb} & n/d & - & $\medbullet$ & & & & FedAvg & $\medbullet$ & $\medbullet$ & $\medbullet$ & & $\medbullet$\\
    FATS \cite{tao2024communication} & Non-IID & Dirichlet & $\medbullet$ & & $\medbullet$ &  & FedAvg & $\medbullet$ & $\medbullet$ & $\medbullet$ & & $\medbullet$\\
    FedADP \cite{jiang2024efficient} & n/d & - &  $\medbullet$ & & $\medbullet$ &  & FedAvg & $\medbullet$ & & $\medbullet$ & $\medbullet$\\
    Shao et al. \cite{shao2024federated} & Non-IID & Unique &     $\medbullet$ & & &  & Weighted Avg & $\medbullet$ & & $\medbullet$ & & $\medbullet$ \\
    Wang et al. \cite{wang2024forget}  & IID & - &      & & & $\medbullet$ &  FedAvg & $\medbullet$ & $\medbullet$ \\
    FedRecover \cite{cao2023fedrecover} & Non-IID & Fang & $\medbullet$ & $\medbullet$ & & & FedAvg, Med, TrMean & $\medbullet$ & & $\medbullet$ & $\medbullet$ \\
    Wu et al. \cite{wu2024unlearning} & n/d & - & $\medbullet$ & & & & FedAvg & $\medbullet$ & $\medbullet$ \\
    Fu et al. \cite{fu2024client} & IID, Non-IID  & Fang & $\medbullet$ & & & &  n/d & $\medbullet$ & $\medbullet$ & $\medbullet$ \\
    FedRecovery \cite{zhang2023fedrecovery} & IID & - & $\medbullet$ & & &  & FedAvg & $\medbullet$ & $\medbullet$ & $\medbullet$ & $\medbullet$ & $\medbullet$ \\
    CFRU \cite{huynh2025certified} & Non-IID & n/d & & & $\medbullet$ & & FedAvg & $\medbullet$ & $\medbullet$ & & & $\medbullet$ \\
    MetaFul \cite{wang2023mitigating} & IID, Non-IID & Dirichlet & $\medbullet$ & & & $\medbullet$ & FedAvg & $\medbullet$ & $\medbullet$ & $\medbullet$\\
    Deng et al. \cite{deng2023vertical} & IID & - & $\medbullet$ & $\medbullet$ & & & n/d & $\medbullet$ & $\medbullet$ & $\medbullet$ & $\medbullet$\\
    Starfish \cite{liu2024privacy} &  n/d & - & $\medbullet$ & & & & FedAvg & $\medbullet$ & $\medbullet$ & $\medbullet$ & $\medbullet$ & $\medbullet$  \\
    Crab \cite{jiang2024towards}  &  n/d & - & $\medbullet$ & & $\medbullet$ &  & FedAvg & $\medbullet$ & $\medbullet$ & $\medbullet$ \\
    FedEraser \cite{liu2021federaser} & n/d & - & $\medbullet$ & $\medbullet$ & & & FedAvg & $\medbullet$ & $\medbullet$ & $\medbullet$ \\
    FRU \cite{yuan2023federated} &  n/d & - & & & $\medbullet$ & & FedAvg & $\medbullet$ & $\medbullet$ & $\medbullet$ & & & $\medbullet$ & $\medbullet$\\
    SIFU \cite{fraboni2024sifu} & IID, Non-IID & Dirichlet & $\medbullet$ & & &  & FedAvg & $\medbullet$ & $\medbullet$ & $\medbullet$ & $\medbullet$ & $\medbullet$ \\
    FedUnlearn \cite{nguyen2024empirical} &IID & - & $\medbullet$ & & & & FedAvg & $\medbullet$ & $\medbullet$ \\
    FedUNRAN \cite{mora2024fedunran} & Non-IID & Random & $\medbullet$ & & & & FedAvg & $\medbullet$ & $\medbullet$ & $\medbullet$ \\
    FUCRT \cite{guo2024forgetting} & IID, Non-IID & Dirichlet & $\medbullet$ & & & & n/d & $\medbullet$ & $\medbullet$ & $\medbullet$ & $\medbullet$\\
    FedMUA \cite{chen2025fedmua} & IID, Non-IID & Dirichlet & $\medbullet$ & $\medbullet$ & & & FedAvg & $\medbullet$ & $\medbullet$ & \\
    Pan et al. \cite{pan2025feature} & IID, Non-IID & Random & & $\medbullet$ & & & Weighted Avg. & $\medbullet$ & $\medbullet$ & $\medbullet$ & & $\medbullet$\\
    FFMU \cite{che2023fast} & n/d & - & $\medbullet$ & & &  & FedAvg & $\medbullet$ & $\medbullet$ & $\medbullet$ & & $\medbullet$\\
    FedFilter \cite{wang2023edge} &Non-IID & - &  & & $\medbullet$ & & Avg. base layers & $\medbullet$ \\
    UKRL \cite{xu2023revocation} & IID, Non-IID & Random &  $\medbullet$ & & &  & FedAvg & $\medbullet$ & & $\medbullet$ & $\medbullet$  \\
    MoDe \cite{zhao2023federated} & Non-IID & Dirichlet & $\medbullet$ & & &  & FedAvg & $\medbullet$ & $\medbullet$ & $\medbullet$ \\
    ConDa \cite{chundawat2024conda} & IID, Non-IID & Random & $\medbullet$ & & &  & FedAvg & $\medbullet$ & $\medbullet$ & $\medbullet$ \\
    RobustFU \cite{sheng2024robust} & IID, Non-IID & n/d & & $\medbullet$ & &  & n/d & $\medbullet$ & $\medbullet$ & $\medbullet$ & $\medbullet$ & $\medbullet$ \\ 
    FRAMU \cite{shaik2024framu} & Non-IID & Concept drift & $\medbullet$ & $\medbullet$ & $\medbullet$ & $\medbullet$ & FedAvg & $\medbullet$ & $\medbullet$ & $\medbullet$ & & & $\medbullet$ \\
    VeriFi \cite{gao2024verifi} & Non-IID & Dirichlet & $\medbullet$ & & $\medbullet$ &  & FedAvg, Krum, Median & $\medbullet$ & $\medbullet$ & $\medbullet$ & $\medbullet$ \\
    Lin et al. \cite{lin2024blockchain} & n/d & - & $\medbullet$ & & &  & Weighted Avg & $\medbullet$ & $\medbullet$ & $\medbullet$ & $\medbullet$ & & $\medbullet$ \\
    FC \cite{pan2022machine} & IID, Non-IID & n/d &   $\medbullet$ & $\medbullet$ & $\medbullet$ &  & SCMA & $\medbullet$ & $\medbullet$ & $\medbullet$ & $\medbullet$ & $\medbullet$\\ 
    Wang et al. \cite{wang2022federated} & IID, Non-IID & Fang & $\medbullet$ & & &  & FedAvg & $\medbullet$ & $\medbullet$ & $\medbullet$ \\ 
    FAST \cite{guo2023fast} & IID, Non-IID & Random &  $\medbullet$ & & &  & FedAvg & $\medbullet$ & $\medbullet$ & $\medbullet$ & $\medbullet$ \\
    ElBedoui et al. \cite{elbedoui2023ecg} & IID & - & & & & $\medbullet$  & FedAvg & $\medbullet$  \\
    Fast-FedUL \cite{huynh2024fast} & Non-IID & Dirichlet & $\medbullet$ & & & & SecAgg & $\medbullet$ & $\medbullet$ & $\medbullet$ & & $\medbullet$ \\
    NoT \cite{khalil2025not} & IID, Non-IID & Dirichlet & $\medbullet$ & & & & FedAvg & $\medbullet$ & $\medbullet$ & $\medbullet$& \\
    2F2L \cite{jin2023forgettable} & IID & - & $\medbullet$ & & &  & FedAvg & $\medbullet$ & $\medbullet$ \\
    Liu et al. \cite{liu2022right} & IID & - &  $\medbullet$ & & & & FedAvg & $\medbullet$ & $\medbullet$ & $\medbullet$ & & $\medbullet$ & $\medbullet$\\
    Li et al. \cite{li2024federated} &  n/d & - & $\medbullet$ & & & & FedAvg & $\medbullet$ & $\medbullet$ & & & & & $\medbullet$ \\
    BadUnlearn \cite{wang2025poisoning} & Non-IID & Dirchlet & $\medbullet$ & & & & FedAvg, Med, TrMean & $\medbullet$ & $\medbullet$ & & $\medbullet$ & $\medbullet$  \\
    FedU \cite{wang2024fedu} & IID & - & $\medbullet$ & & & & n/d & $\medbullet$ & $\medbullet$ & $\medbullet$ & & $\medbullet$  \\
    FedME2 \cite{xia2023fedme} & n/d & - & $\medbullet$ & & & & FedAvg & $\medbullet$ & $\medbullet$ & & $\medbullet$  \\
    Alam et al. \cite{alam2023get} & IID & - &  $\medbullet$ & & & &  FedAvg & $\medbullet$ & $\medbullet$ \\
    BFU \cite{wang2023bfu} & n/d & - & $\medbullet$ & & & &  FedAvg & $\medbullet$ & $\medbullet$ & $\medbullet$\\
    FedHarmony \cite{dinsdale2022fedharmony} & Non-IID & Covariate shift & $\medbullet$ & & &  & FedEqual & $\medbullet$ \\    
    Chen et al. \cite{chen2024federated} & IID, Non-IID & n/d & $\medbullet$ & & & $\medbullet$ & n/d & $\medbullet$ & $\medbullet$ & $\medbullet$  \\
    FCU \cite{deng2024enable} & Non-IID & Dirichlet & $\medbullet$ & & & & FedAvg & $\medbullet$ & $\medbullet$ & $\medbullet$& \\
    Ferrari \cite{gu2024ferrari} & IID, Non-IID & Dirichlet &  &  & & $\medbullet$ & n/d & $\medbullet$ & $\medbullet$ & $\medbullet$ & & $\medbullet$ \\ 
    Imba-ULRc \cite{yu2024federated} & Non-IID & Random & $\medbullet$ & & & & Weighted Avg & $\medbullet$ & $\medbullet$ & & & & $\medbullet$ \\
    Xu et al. \cite{xu2024update} & IID, Non-IID & n/d & $\medbullet$ & & & & FedAvg & $\medbullet$ & $\medbullet$ & $\medbullet$& \\
    TrustChain \cite{zuo2024federated} & Non-IID & n/d & & & $\medbullet$ & & n/d & $\medbullet$ & & $\medbullet$ \\
    FUSED \cite{Zhong2025UnlearningTK} & Non-IID & Dirichlet & $\medbullet$ & &  & & FedAvg & $\medbullet$ & $\medbullet$ & $\medbullet$ & $\medbullet$ & \\
    Goldfish \cite{wang2024goldfish} & Non-IID & Random & $\medbullet$ & & &  & n/d & $\medbullet$ & $\medbullet$ & $\medbullet$ & & & $\medbullet$  \\
    FedLU \cite{zhu2023heterogeneous} & Non-IID & Unique & & & & $\medbullet$  & FedAvg & $\medbullet$ & $\medbullet$ & $\medbullet$ \\
    FedAF \cite{li2023federated} & n/d & - & $\medbullet$ & & &  & FedAvg & $\medbullet$ & $\medbullet$ & $\medbullet$ & $\medbullet$\\
    FedQUIT \cite{mora2024fedquit} & Non-IID & Dirichlet & $\medbullet$ & & & & n/d & $\medbullet$ & $\medbullet$ & \\
    CKGD  \cite{zhang2025model} & IID, Non-IID & Dirichlet & $\medbullet$ & & & & FedAvg & $\medbullet$ & $\medbullet$ & $\medbullet$& \\
    SFU \cite{zhou2024streamlined} & IID, Non-IID & Random & $\medbullet$ & & $\medbullet$ & & FedAvg & $\medbullet$ & $\medbullet$ & $\medbullet$ & & $\medbullet$ \\
    HDUS \cite{ye2023heterogeneous} & Non-IID & Unique & $\medbullet$ & & & &  n/d & $\medbullet$ & & & & & $\medbullet$ & $\medbullet$ \\
    EWC-SGA \cite{wu2022federated} & IID, Non-IID & Unique & $\medbullet$ & & & &  FedAvg & $\medbullet$ & $\medbullet$ \\
    SFU \cite{li2023subspace} & IID, Non-IID & Dirichlet & $\medbullet$ & & &  & n/d & $\medbullet$ & $\medbullet$ & & $\medbullet$ \\
    Halimi et al. \cite{halimi2022federated} & IID & - & $\medbullet$ & & &  & FedAvg & $\medbullet$ & $\medbullet$ & $\medbullet$\\
    QuickDrop \cite{dhasade2024quickdrop} & IID, Non-IID & Dirichlet & $\medbullet$ & & &  & FedAvg & $\medbullet$ & $\medbullet$ & $\medbullet$ & & & & $\medbullet$ \\
    LMR \cite{ameen2024addressing} & Non-IID & n/d & $\medbullet$ & & & & n/d & $\medbullet$ & $\medbullet$ & $\medbullet$ & & & $\medbullet$ \\
    Wang et al. \cite{wang2024server} & Non-IID & n/d & $\medbullet$ & & & & FedAvg & $\medbullet$ & $\medbullet$ & $\medbullet$ & \\
    Han et al. \cite{han2025vertical} &  n/d & - & $\medbullet$ & & & & FedAvg & $\medbullet$ & $\medbullet$ & $\medbullet$ & \\
    FedOSD \cite{pan2024federated} & IID, Non-IID & Unique & $\medbullet$ & & & & FedAvg& $\medbullet$ & $\medbullet$ & $\medbullet$ & & $\medbullet$ \\
    F2ul \cite{su2024f2ul} & IID, Non-IID & n/d & $\medbullet$ & & & & FedAvg & $\medbullet$ & $\medbullet$ & $\medbullet$ & & & $\medbullet$ \\
    VFU \cite{varshney2025unlearningclientsfeaturessamples} & n/d & - & $\medbullet$ & $\medbullet$ & & & n/d & $\medbullet$ & $\medbullet$ & $\medbullet$& \\
    forgetSVGD \cite{gong2022forget} & Non-IID & Unique & $\medbullet$ & & & & n/d & $\medbullet$ & $\medbullet$ \\
    CforgetSVGD \cite{gong2022compressed} & Non-IID & Unique & $\medbullet$ & & & &  FedAvg & $\medbullet$ & $\medbullet$ & $\medbullet$ \\
    KNOT \cite{su2023asynchronous} & Non-IID & Dirichlet & $\medbullet$ & $\medbullet$ & $\medbullet$ & &  FedAvg, FedBuff & $\medbullet$ & $\medbullet$ & $\medbullet$ & & & $\medbullet$ & $\medbullet$ \\
    Lin et al. \cite{lin2024scalable} & IID, Non-IID & Random & $\medbullet$ & & $\medbullet$ & &  FedAvg & $\medbullet$ & $\medbullet$ & $\medbullet$ & & $\medbullet$ & $\medbullet$ & $\medbullet$ \\
    Lin et al. \cite{lin2024incentive} & IID, Non-IID & n/d & $\medbullet$ & & & &  FedAvg & $\medbullet$ & $\medbullet$ \\
    Liu et al.  \cite{liu2024guaranteeing} & n/d & - & $\medbullet$ & & & & SecAgg & $\medbullet$ & $\medbullet$ & $\medbullet$ & $\medbullet$ & $\medbullet$& \\
    Hier-FUN \cite{ma2024hier} & Non-IID & Random & $\medbullet$ & & & & FedAvg & $\medbullet$ & $\medbullet$ & $\medbullet$ & & $\medbullet$ & & $\medbullet$ \\
    FedUMP \cite{zhu2024federated} & n/d & - & $\medbullet$ & & & & Selective Avg. & $\medbullet$ & $\medbullet$ & $\medbullet$ & & & & $\medbullet$ \\
    k-IPfedAvg \cite{varshney2024efficient} & IID, Non-IID & n/d & $\medbullet$ & & & & FedAvg, k-IPfedAvg & $\medbullet$ & & $\medbullet$ \\
    FedAU \cite{gu2024unlearning} & IID, Non-IID & Dirichlet & $\medbullet$ & & & & n/d & $\medbullet$ & $\medbullet$ & $\medbullet$ & & $\medbullet$\\
    \bottomrule
    \end{tabular}
\end{table*}

\subsection{On What Dataset?}

\begin{figure}
\centerline{\includegraphics[width=1\linewidth]{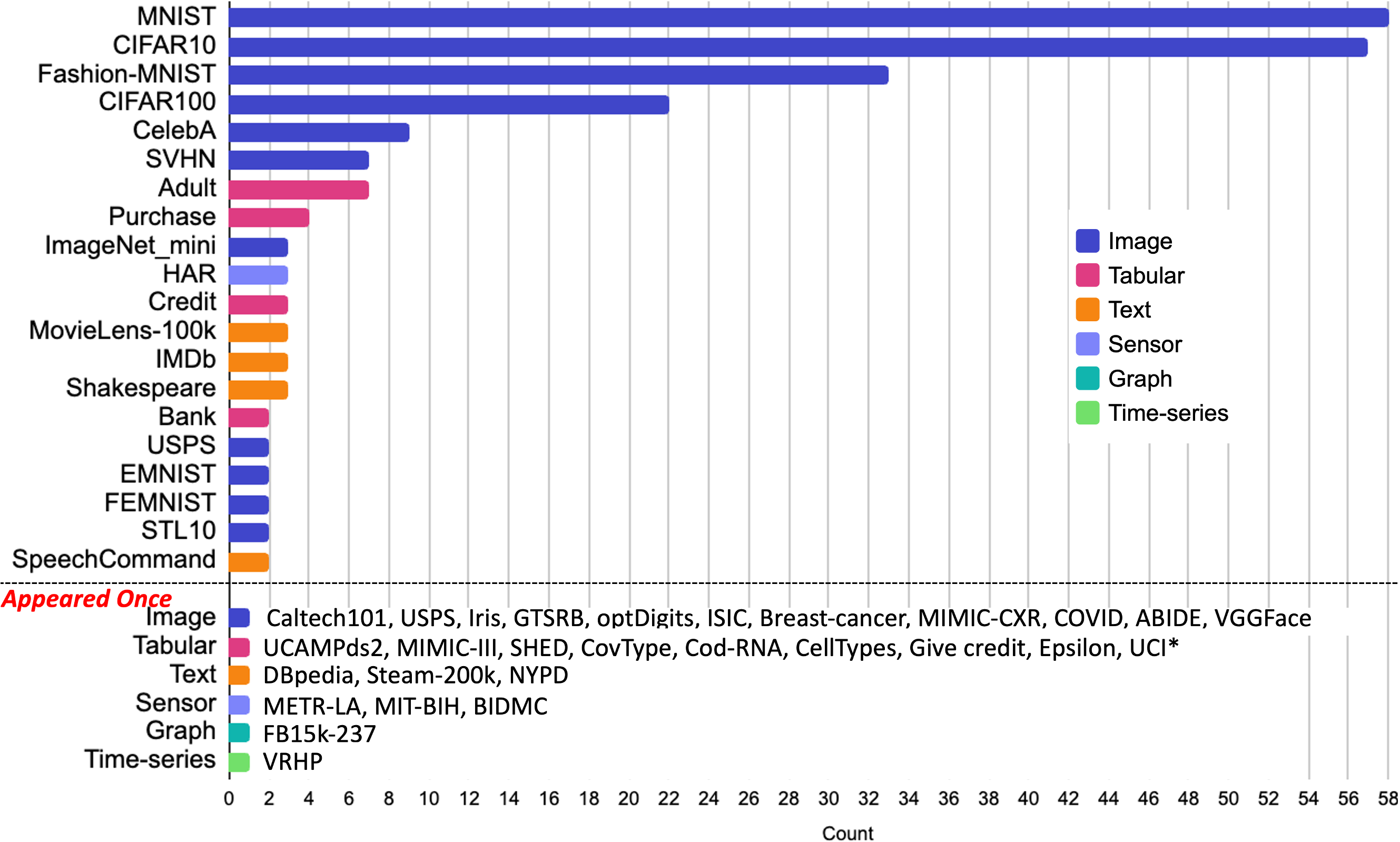}}
\caption{Dataset choice for experiments.}
\label{fig:dataset}
\vspace{-10pt}
\end{figure}

Existing FU methods demonstrated their effectiveness in diverse types of training data and for various tasks, such as image classification, object detection, regression, sentiment analysis, recommendation, and clustering.
Nonetheless, as depicted in \autoref{fig:dataset}, about 53 out of 94 studies demonstrated its effectiveness primarily on simple image datasets like MNIST \footnote{https://yann.lecun.com/exdb/mnist/} and CIFAR10 \footnote{https://www.cs.toronto.edu/~kriz/cifar.html} while fewer works have experimented on more complex image datasets such as CelebA \footnote{https://mmlab.ie.cuhk.edu.hk/projects/CelebA.html}, CIFAR100 or EMNIST\footnote{https://www.nist.gov/itl/products-and-services/emnist-dataset}.
In total, image datasets were used 209 times, 80\% of the entire usage, and multi-modal datasets\footnote{TCGA https://www.cancer.gov/ccg/research/genome-sequencing/tcga \\ TMI https://opencas.webarchiv.kit.edu/?q=tmidataset} were used twice in a single work, \cite{pan2022machine}. 

As a vast number of works have proven their unlearning capabilities with simple image datasets, recent works are expanding their experiments with either more complex images or text or both datasets.
In particular, mostly from 2024, proposed FU works demonstrated their capabilities with image and text datasets (e.g., CIFAR100 and Shakespear datasets), showing their method is not limited to a specific type of dataset \cite{su2023asynchronous, gao2024verifi, tao2024communication, lin2024scalable, wang2024efficient, jiang2024efficient, wu2024unlearning, gu2024ferrari, zhou2024streamlined}.

\begin{tcolorbox}[enhanced, breakable, fontupper=\linespread{.93}\selectfont]    \textbf{\hypertarget{t4}{T-4.} Takeaways on Dataset Usage.}
    Like the early stages of FL, FU is yet predominantly focused on vision tasks with simple image datasets. Nonetheless, the practicality of FU should not be limited to basic datasets like MNIST. For FU to fully leverage the advantages offered by AI and FL, it should extend its scope to include diverse datasets for more sensitive tasks beyond vision tasks on images.    
\end{tcolorbox}

\subsection{Configurations}

As detailed in \autoref{tab:settings}, images are the most used data type for vision tasks in FU, so convolution networks were employed the most. 
Model architectures vary from shallower designs (2- to 4-layer CNN) to more complex structures, including LeNet, VGG, ResNet, DenseNet, and MobileNet. Most recent works in 2025 \cite{khalil2025not, wang2025unlearning} include vision transformers ViT-B/16 and SimpleViT, respectively. 
For some language tasks, we can observe the use of LSTM in \cite{zhou2024streamlined, tao2024communication}, BERT in \cite{gu2024ferrari}, and GPTs in \cite{zuo2024federated, su2023asynchronous, lin2024scalable}.
In a different domain, MetaFul \cite{wang2023mitigating} utilized LSTM to train a model on the video dataset. 
FRU \cite{yuan2023federated} and CFRU \cite{huynh2025certified}, a federated unlearning method in recommender systems, opted for graph-based models like NCF and LightGCN.
For Bayesian FU implementations, as seen in works like \cite{wang2023bfu, gong2022forget}, the Bayesian Neural Network (BNN) architecture was used. \cite{zhu2023heterogeneous} employed knowledge graph embedding models (TransE, ComplEx, and RotE) specifically tailored for graph-type datasets. 
A detailed list of model architecture is described in \autoref{app:comprehensive}.

To address asynchronous FU, FedHarmony \cite{dinsdale2022fedharmony} introduced FedBuff for integrating client updates. 
Previous works had been primarily focused on unlearning in an HFL setting, more recently, from 2023, unlearning was explored in various federated learning settings, such as VFL/cross-silo FL \cite{han2025vertical, deng2023vertical, zhang2023securecut, wang2024efficient, varshney2025unlearningclientsfeaturessamples} and blockchain-based FL \cite{liu2024blockful, lin2024blockchain}.

The prevailing aggregation method in FU remains \texttt{FedAvg}, weighing each client's gradient based on their dataset size and averaging them to ensure proportional impact. Some work \cite{cao2023fedrecover, gao2024verifi, wang2025poisoning} involved Byzantine-robust aggregation algorithms--Krum \cite{blanchard2017machine}, Trimmed-Mean \cite{yin2018byzantine}, and Median \cite{yin2018byzantine} to protect the learning system from backdoor or poisoning attacks. 
These aggregation rules were used to preserve the security or integrity of the model, effectively handling outliers by excluding the most deviating updates \cite{blanchard2017machine}, trimming a specific percentage of extreme updates, or computing the coordinate-wise median \cite{yin2018byzantine}.


\subsection{Research Implication of Prior Works}

As summarized in \autoref{tab:settings}, researchers primarily consider efficacy, fidelity, and efficiency when designing FU methods. However, due to the distributed nature of FL and the vulnerabilities that arise in FU, they have also sought theoretical guarantees for convergence \cite{xia2023fedme, liu2024guaranteeing, fraboni2024sifu, wu2022federated} and performance \cite{zhang2023fedrecovery, che2023fast, pan2022machine, liu2021revfrf, huynh2024fast, zhou2024streamlined, jiang2024feduhb, pan2024federated, wang2024fedu}.
Other key concerns include scalability with respect to the number of clients \cite{ye2023heterogeneous, su2023asynchronous, dhasade2024quickdrop, zhu2024federated, yuan2023federated, ma2024hier}, adaptivity to changing data distributions \cite{shaik2024framu, pan2025feature, wang2025forgettingdatatimetheoretically}, asynchronous parameter update timing \cite{su2023asynchronous, wang2025forgettingdatatimetheoretically}, compatibility with different model architectures \cite{ye2023heterogeneous, liu2022right, yuan2023federated, xiong2024appro, wang2025forgettingdatatimetheoretically}, and security against data pollution or leakage, including backdoor and poisoning attacks \cite{pan2022machine, liu2021revfrf, cao2023fedrecover, deng2023vertical, fraboni2024sifu, gao2024verifi, wang2025poisoning, zhang2023securecut, zhang2023fedrecovery, alam2023get, li2023subspace, xu2023revocation}.
Recent works \cite{sheng2024robust, shao2024federated, su2024f2ul, meerza2024confuse, yu2024federated, lin2024incentive, ding2023incentive} have also addressed fairness to prevent unbalanced unlearning \cite{lin2024incentive} and encourage meaningful client participation \cite{su2024f2ul}.

Notably, while efficacy, fidelity, and efficiency are empirically validated, security, adaptivity, and scalability are often only theoretically guaranteed in the respective papers. Security mechanisms primarily rely on Homomorphic Encryption \cite{deng2023vertical, zhang2023securecut}, Differential Privacy \cite{zhang2023fedrecovery, alam2023get, li2023subspace, xiong2024appro}, or trusted third parties \cite{xu2023revocation} rather than formal security analysis—RevFRF \cite{liu2021revfrf} being a notable exception.
Although recent blockchain-based approaches \cite{liu2024blockful, lin2024blockchain, zuo2024federated, zuo2025federated} aim to enhance scalability and security through their inherent framework design, they still lack strong theoretical guarantees in these aspects.

\begin{tcolorbox}[enhanced, breakable, fontupper=\linespread{.93}\selectfont]    
\textbf{\hypertarget{t5}{T-5.} Takeaways on Configurations and Implications.}
    Most works aim to achieve efficacy, fidelity, or efficiency, with a few additional focuses on security, guarantee, scalability, and adaptability. 
    For instance, over 90\% of them relied on the simplest \texttt{FedAvg}, although it could lead to intolerance to Byzantine failures or heterogeneity. 
    Given increased vulnerabilities due to accessibility to both learned and unlearned models and dynamic client participation, it becomes imperative to consider the other implications actively.
\end{tcolorbox}

\section{Existing Unlearning Techniques}
\label{sec:technique}

\begin{table*}[htbp]
    \captionsetup{skip=4pt}
    \caption{Unlearning targets, influence removal methods, and performance recovery methods define an unlearning mechanism. Each refers to what we want the model to forget, how the model can forget, and how to maintain the performance on the remembered dataset, respectively. \textbf{sp, cs, ct}, and \textbf{ft} refer to sample, class, client, and feature, respectively. } 
    \label{tab:unlearning_mechanism}
    \footnotesize
    \renewcommand*{\arraystretch}{0.9}
    \setlength\tabcolsep{4.6pt} 
    \begin{tabular}{R{2.3cm}C{0.01cm}C{0.01cm}C{0.01cm}C{0.01cm}>{\setlength{\baselineskip}{0.8\baselineskip}}L{7.4cm}>{\setlength{\baselineskip}{0.8\baselineskip}}L{6.2cm}}
    \toprule
    \multicolumn{1}{c}{\multirow{2}*{\textbf{Ref.}}} & \multicolumn{4}{c}{\textbf{Target}} & \multicolumn{2}{c}{\textbf{Unlearning Method}}  \\  \cmidrule{2-7}
    & \textbf{sp} & \textbf{cs} & \textbf{ct} & \textbf{ft} & \multicolumn{1}{c}{\textbf{Influence removal}} & \multicolumn{1}{c}{\textbf{Performance recovery}} \\ \toprule

    RevFRF\cite{liu2021revfrf} & & & $\medbullet$ &	
        & Remove the node of the target client and all the child nodes	& - \\
    Exact-Fun\cite{xiong2023exact} & $\medbullet$ & & &	
        & Retrain the remaining clients using a quantized model	& - \\
    FedUHB\cite{jiang2024feduhb} & & & $\medbullet$ &	
        & Rapid retraining: accelerating using heavy ball method	& - \\
    FATS\cite{tao2024communication} & $\medbullet$ & & $\medbullet$ &	
        & Retrain only if sampling probability changes after removal	 & - \\
    R2S\cite{wang2024efficient} & & & & $\medbullet$	
        & Retrain from historically saved local model checkpoints	& Automatic optimizer control for generalization\\
    Zuo et al.\cite{zuo2024federated} & $\medbullet$ & $\medbullet$ & $\medbullet$ & 	
        & Rollback before the client joins, retraining in remaining clients	& - \\
    FedADP\cite{jiang2024efficient} & & & $\medbullet$ &	
        & Recover from historical local updates based on cosine sim.\footnotemark	 & - \\
    Shao et al.\cite{shao2024federated} & & & $\medbullet$ &	
        & Recover from historically saved local updates	& Add a penalty term derived from projected gradients \\
    Wang et al.\cite{wang2024forget}  & & & $\medbullet$ &	
        & Recover from historically saved local model updates  & - \\ 
    FedRecover\cite{cao2023fedrecover} & & & $\medbullet$ &	
        & Recover from historically saved remaining clients' updates	& Pre- and Post-training \\
    Wu et al.\cite{wu2024unlearning} & $\medbullet$ & & &	
        & Recover from historically saved remaining clients' updates	& KD: transfer remembered knowledge \\
    Fu et al.\cite{fu2024client} & & & $\medbullet$ &	
        & Recover from historically saved local updates by projection	& Post KD iterations on auxiliary data in the server \\
    FedRecovery\cite{zhang2023fedrecovery} & & & $\medbullet$ &	
        & Recover from historically saved local model updates	& Add Gaussian noise \\
    CFRU\cite{huynh2025certified} & & & $\medbullet$ &	
        & Recover from sampled historical updates	& - \\
    FedADP\cite{wang2024efficient} & & & $\medbullet$ & 	
        & Recover from selectively stored historical info & Post-training with remaining clients \\ 
    MetaFul\cite{wang2023mitigating} & & & $\medbullet$ &	
        & Recover from saved updates and subtract target ones	& Fine-tuning using the direction of the target updates  \\
    Deng et al.\cite{deng2023vertical} & & & $\medbullet$ & $\medbullet$	
        & Recover from saved target model updates at the last round by subtracting them from the global model	& Constraints on intermediate local model parameter during training \\
    Starfish\cite{liu2024privacy} & & & $\medbullet$ &	
        & Recover from least-affected local updates	& Recompute the local updates with large approx error \\
    Crab\cite{jiang2024towards}  & & & $\medbullet$ &	
        & Recover from selectively saved least-affected local updates	& - \\  
    FedEraser\cite{liu2021federaser} & & & $\medbullet$&	
        & Recover from periodically saved local model updates	& Post-training: a few training rounds after unlearning \\
    FRU\cite{yuan2023federated} &  & & $\medbullet$ &	
        & Restore certain global model updates saved historically	& Gradient manipulation: direction \\
    SIFU\cite{fraboni2024sifu} & & & $\medbullet$ &	
        & Restore and perturb certain global models, saved historically	& - \\
    FedRemover\cite{yuan2024towards} & & & $\medbullet$ &	
        & Adjust direction from historically saved local updates only if the model performance drops	& - \\
    ConFUSE\cite{meerza2024confuse} & $\medbullet$ & & $\medbullet$ & $\medbullet$	
        & Create a fake dataset to confuse the model	& Update salient weights only \\    
    Imba-ULRc\cite{yu2024federated} & & & $\medbullet$ &	
        & Finetuning on oversampled, noise-removed data 	& Post-training with remaining clients \\
    FedUnlearn\cite{nguyen2024empirical} & & & $\medbullet$ &	
        & Perturb target label to the fake one	& - \\
    FedUNRAN\cite{mora2024fedunran} & & & $\medbullet$ & 	
        & Perturb target label to the random ones	& Post-training with remaining clients \\
    FUCRT\cite{guo2024forgetting} & & $\medbullet$ & &	
        & Perturb target label to the random ones	& Optimize local space to be aligned with global space \\
    FedMUA\cite{chen2025fedmua} & & & $\medbullet$ &	
        & Modify target sample prediction	& Identify the most influential samples \\
    Pan et al.\cite{pan2025feature} & & & & $\medbullet$	
        & Perturb target client updates	& Gradient alignment \\
    SecForget\cite{liu2020learn} & $\medbullet$ & & &	
        & Perturbation addition to the target client	& Add a regularization term: L1 norm \\
    FFMU\cite{che2023fast} & $\medbullet$ & & &	
        & Perturbation addition to the target client	& - \\
    FedFilter\cite{wang2023edge} & $\medbullet$ & & &	
        & Perturbing target update by generating a reverse gradient	& SGD to minimize impact on the model accuracy \\
    UKRL\cite{xu2023revocation} & & & $\medbullet$ &	
        & Perturbing target updates by training it on noised input	& - \\
    MoDe\cite{zhao2023federated} & & $\medbullet$ & $\medbullet$ &	
        & Scale down on the target data points	& Guided fine-tuning on remaining data points \\
    ConDa\cite{chundawat2024conda} & & & $\medbullet$ &	
        & Scale down the global param most impacted by target clients	& - \\
    RobustFU\cite{sheng2024robust} & & $\medbullet$ & &	
        & Scale up/down based on label prediction discrepancies	& Introduce random samples based on the discrepancy\\
    FRAMU\cite{shaik2024framu} & $\medbullet$ & & &	
        & Scale up/down using an attention mechanism	& - \\
    VeriFi\cite{gao2024verifi} & & & $\medbullet$ &	
        & Scale up/down on remaining/target clients' updates	& - \\
    Lin et al.\cite{lin2024blockchain} & & & $\medbullet$ &	
        & Scale up/down on remaining clients' updates & - \\ 
    FC\cite{pan2022machine} & $\medbullet$ & & $\medbullet$ &	
        & Scale to zero on the target clients' updates	& - \\
    Wang et al.\cite{wang2022federated} & & $\medbullet$ & &	
        & Prune the channel of gradients corresponding to the target	& Fine-tuning: few FL training without regularization \\
    Xu et al.\cite{xu2024update} & & $\medbullet$ & &	
        & Prune less influential channel in identifying target	& - \\
    FUSED\cite{Zhong2025UnlearningTK} & $\medbullet$ & $\medbullet$ & $\medbullet$ & 
        & Randomly dropout parameters in critical layers & -  \\ 
    SecureCut\cite{zhang2023securecut} & $\medbullet$ & & & $\medbullet$	
        & Prune the nodes of their local model and retrain it	& Post-training with remaining clients \\
    FAST\cite{guo2023fast} & & & $\medbullet$ &	
        & Subtract target model updates from the global model	& Fine-tuning with a small-size IID dataset \\
    ElBedoui et al.\cite{elbedoui2023ecg} & $\medbullet$ & & $\medbullet$ &	
        & Subtract target model updates from the global model \footnotemark	& - \\
    Fast-FedUL\cite{huynh2024fast} & & & $\medbullet$ &	
        & Subtract differences between target model updates and the aggregation of selectively stored local updates	& - \\
    NoT\cite{khalil2025not} & & & $\medbullet$ &	
        & Subtract targets' layer-wise parameters from the global model	& Post-training with remaining clients \\
    Wang et al.\cite{wang2025forgettingdatatimetheoretically} & $\medbullet$ & & & $\medbullet$	
        & Subtract target confidence from a confidence matrix	& - \\
    2F2L\cite{jin2023forgettable} & & & $\medbullet$ &	
        & Loss function approximation	& - \\
    Liu et al.\cite{liu2022right} & & & $\medbullet$&	
        & Loss function approximation using diagonal FIM	& Apply Momentum technique to Hessian diagonal \\
    Xie et al.\cite{xie2024adaptive}	& & $\medbullet$ & &
        & Loss function approximation using FIM	& Parameter clipping \& KD using synthesized samples \\
    Li et al.\cite{li2024federated} & & & $\medbullet$ &	
        & Compute approximate Hessian to estimate recovered gradient	& Gradient clipping \\
    BadUnlearn\cite{wang2025poisoning} & & & $\medbullet$ &	
        & Compute approximate Hessian using limited historical data	& - \\
    Appro-Fun\cite{xiong2024appro} & & & $\medbullet$ &	
        & Compute approximate Hessian	& Post-training with remaining clients \\
    FedU\cite{wang2024fedu} & & & $\medbullet$ &	
        & Approximate influence using Hessian-vector product and subtract it from the global model	& local post-training with remaining dataset \\
    FedME2\cite{xia2023fedme} & & & & $\medbullet$	
        & Multi-task learning: data erasure and remembrance	& Add a regularization term: L2 norm \\
    Alam et al.\cite{alam2023get} & $\medbullet$ & & &	
        & Multi-task learning: data erasure and remembrance	& Dynamic penalization \\
    BFU\cite{wang2023bfu} & $\medbullet$ & & &	
        & Multi-task learning: data erasure	& Multi-task learning: maintaining performance \\
    FedHarmony\cite{dinsdale2022fedharmony} & & & &$\medbullet$	
        & Multi-task learning: data erasure and remembrance	& - \\
    Chen et al.\cite{chen2024federated} & $\medbullet$ & & &	
        & Multi-task learning: minimize KLD between the predicted distribution of forget and arbitrary data	& Multiply the loss by a scaling factor to balance between efficacy and fidelity \\
    Goldfish\cite{wang2024goldfish} & $\medbullet$ & & &	
        & KD \& Multi-task learning: minimize prediction difference within the target set and transfer only retaining knowledge	& Multi-task learning: maintaining performance \\
    FCU\cite{deng2024enable} & & & $\medbullet$ &	
        & Contrastive unlearning to make it similar to a model trained only on the retaining set at feature level	& Post-training on low-frequency components \\
    FedLU\cite{zhu2023heterogeneous} & $\medbullet$ & & &	
        & KD: transfer knowledge to be remembered	& Fine-tuning: suppressing activation of the target data \\
    FedAF\cite{li2023federated} & $\medbullet$ & $\medbullet$ & $\medbullet$ &	
        & KD: transfer fake knowledge	& EWC training, using Hessian matrix as a regularizer \\
    FedQUIT\cite{mora2024fedquit} & & & $\medbullet$ &	
        & KD: transfer random prediction on forget data	& KD: transfer retaining knowledge \\
    CKGD \cite{zhang2025model} & & & $\medbullet$ &	
        & KD: clip-guided few-shot kd from historically saved updates	& - \\
    SFU\cite{zhou2024streamlined} & $\medbullet$ & & &	
        & KD: negative transfer knowledge to be forgotten & KD: transfer retaining knowledge \\
    HDUS\cite{ye2023heterogeneous} & & & $\medbullet$ &	
        & Ensemble remaining clients' knowledge-distilled seed models	& - \\
    VFU\cite{varshney2025unlearningclientsfeaturessamples} & $\medbullet$ & & $\medbullet$ & $\medbullet$	
        & KD to unlearn feature/clients, GA to unlearn sample & - \\
    \midrule
    \end{tabular}
\end{table*}

\begin{table*}[ht]
    \captionsetup{skip=4pt}
    \label{tab:unlearning_mechanism_cont}
    \footnotesize
    \renewcommand*{\arraystretch}{0.9}
    \setlength\tabcolsep{4.6pt} 
    \begin{tabular}{R{2.3cm}C{0.01cm}C{0.01cm}C{0.01cm}C{0.01cm}>{\setlength{\baselineskip}{0.8\baselineskip}}L{7.4cm}>{\setlength{\baselineskip}{0.8\baselineskip}}L{6.2cm}}
    \midrule
    EWC-SGA\cite{wu2022federated}&$\medbullet$&$\medbullet$&$\medbullet$&	
        & Reverse training: Stochastic Gradient Ascent (SGA)	& EWC training, using FIM as a regularizer \\
    SFU\cite{li2023subspace} & $\medbullet$ & & &	& Reverse training: Gradient Ascent (GA)	& Project the target updates into the subspace\\
    Halimi et al.\cite{halimi2022federated} & & & $\medbullet$ &	& Reverse training: maximizing local loss	& Post-training: a few training rounds after unlearning \\
    QuickDrop\cite{dhasade2024quickdrop} & & $\medbullet$ & $\medbullet$ &	& Reverse training (SGA) on the distilled dataset	& Fine-tuning on augmented remaining data samples \\
    LMR\cite{ameen2024addressing} & $\medbullet$ & & &
        & Reverse training: layer-wise GA	& Post training \\
    Wang et al.\cite{wang2024server} & $\medbullet$ & & &	
        & Reverse training: GA	& Post-training \\
    Han et al.\cite{han2025vertical} & & & $\medbullet$ &	
        & Reverse training: GA guided by avg'd remaining clients	& Post-training with remaining clients \\
    BlockFUL\cite{liu2024blockful} & & $\medbullet$ & $\medbullet$ & 	& Reverse training: GA	& - \\
    FedOSD\cite{pan2024federated} & & & $\medbullet$ &	& Reverse training: GA to orthogonal steepest descent direction	& Project gradient to the original plane \\
    F2ul\cite{su2024f2ul} & & & $\medbullet$ &	& Reverse training: GA	& post training \\
    forgetSVGD\cite{gong2022forget} & $\medbullet$ & & &	& Variational inference, maximizing local loss	& Post-training with remaining clients \\
    CforgetSVGD\cite{gong2022compressed} &$\medbullet$&$\medbullet$&$\medbullet$&	& Variational inference, quantization and sparsification method	& Post-training with remaining clients \\
    KNOT\cite{su2023asynchronous} & & & $\medbullet$ &	& Cluster based on their similarity and retrain one cluster	& - \\
    Lin et al.\cite{lin2024scalable} & & & $\medbullet$ &	& Clustering: retrain the cluster containing the target & - \\ 
    Lin et al.\cite{lin2024incentive} & & & $\medbullet$ &	& Cluster based on their similarity and calibrate the gradient by step length and direction	& Dynamically select remaining clients possessing unbalanced local data \\
    Liu et al.\cite{liu2024guaranteeing} & & & $\medbullet$ &	& Clustering: retrain low performing cluster without target client	& - \\
    Hier-FUN\cite{ma2024hier} & & & $\medbullet$ &	& Clustering and GA on target device	& Post training with remaining clients \\
    FedUMP\cite{zhu2024federated} & & & $\medbullet$ &	& Clustering client\footnotemark and remove the subset model	& - \\
    k-IPfedAvg\cite{varshney2024efficient} & & $\medbullet$ & &	& Remove the target clients from the cluster	& Retrain when any cluster has $\leq$ 2/3/4 \# of clients \\
    FedAU\cite{gu2024unlearning} & $\medbullet$ & $\medbullet$ & $\medbullet$ &	& Add auxiliary unlearning module while training and XOR with the local model upon unlearning request	& - \\
    Ferrari\cite{gu2024ferrari} & & & & $\medbullet$	
        & Guided optimization to minimize feature sensitivity	& - \\
    TrustChain\cite{zuo2025federated} & & & $\medbullet$ &	
        & Finetuning using LoRA	& - \\
    \bottomrule
    \multicolumn{7}{l}{\footnotesize \textsuperscript{10} Cosine Similarity. \textsuperscript{11} Subtracting model updates calculated only on the target data. \textsuperscript{12} based on their distribution}  
    \end{tabular}
    \vspace{-10pt}
\end{table*}

Unlearning methods are classified into exact and approximate. Exact unlearning ensures exact indistinguishability of the distributions between unlearned and retrained models, while approximate unlearning guarantees only approximately \cite{xu2023machine}, often due to imprecise estimation of data influence \cite{li2023federated}.

Despite its precision, exact unlearning comes with substantial overhead compared to approximate one. It necessitates the computation and storage of multiple sub-models during initial training, which can also hinder its adaptability to dynamically changing data \cite{xu2023machine}. Thus, the goal of exact unlearning is retraining as fast as possible with minimal overhead \cite{xiong2023exact, jiang2024feduhb}. Additionally, it is constrained to simpler models due to scalability reasons \cite{schelter2019amnesia} because it often relies on partition-aggregation frameworks \cite{li2023federated}. 

In contrast, approximate unlearning methods offer greater time efficiency gains; for instance, the computational burden is reduced through calculations on sampled parameters instead of computing over all parameters \cite{liu2022right, jin2023forgettable}. Also, model utility can be restored during the performance recovery process. \autoref{tab:unlearning_mechanism} summarizes the unlearning target, influence removal, and performance recovery method for each paper.

\subsection{Influence Removal} 
This stage eliminates the unlearning targets' influences from the trained model, such that the unlearned model behaves as if it has never seen the target data. 
The gray boxes summarize the concepts and limitations. 

\subsubsection{Historical Information} 
The server stores historical local model updates for potential model restoration or estimation before engaging in target unlearning.
In SIFU \cite{fraboni2024sifu}, the server identified and restored an optimal FL iteration without the target information, similar to Starfish \cite{liu2024privacy} and R2S \cite{wang2024efficient}. 
Estimating the unlearned model includes simply restoring historically saved remaining clients' updates \cite{shao2024federated, cao2023fedrecover, wu2024unlearning} or applying calibration upon them by approximating the direction of updates to generate an unlearned model by weighted average \cite{wang2024forget}.
Similarly, FedRecovery \cite{zhang2023fedrecovery} subtracted a weighted sum of gradient residuals from the global model using historical local updates, weighing them by clients' contribution in decreasing global loss. 
Meanwhile, some works subtracted target updates from either global \cite{wang2023mitigating, wu2024unlearning, huynh2024fast, khalil2025not} or remaining clients' models \cite{deng2023vertical, ameen2025speed}. 
Interestingly, Fu et al. \cite{fu2024client} also recovered from the historically saved local model updates but projected them to a low-dimensional subspace inspired by that client-specific information is embedded in their higher-dimensional space.

However, storing all updates is often infeasible, especially for those adapting large models or for cross-device FL with millions of clients. Some works, thus, store updates periodically \cite{liu2021federaser} or selectively \cite{yuan2023federated, jiang2024towards, yuan2024towards, huynh2025certified}, to relax the memory overhead. 
FedEraser \cite{liu2021federaser} took a direction calibration approach as in \cite{wang2024forget} but only using recent sets of updates.
Crab \cite{jiang2024towards} and FRU \cite{yuan2023federated} used a rollback mechanism, as in DBMS, to restore the state before the target engagement and retraining with remaining clients, similar to Zuo et al. \cite{zuo2024federated} and CFRU \cite{huynh2025certified}. 
FRU \cite{yuan2023federated} kept important updates by negative sampling, while Crab \cite{jiang2024towards} only stored high-contributing clients in which the gradients have large KL divergence and high cosine similarity to the global model and identified the least affected gradients to start calibration from.
Monitoring the sampling probability, FATS \cite{tao2024communication} recomputed gradients only when changes occurred after deleting a target to minimize parameter update frequency.

\begin{tcolorbox}[enhanced, breakable, fontupper=\linespread{.93}\selectfont]
    \textbf{\hypertarget{t6}{T-6.} Takeaways on Historical Information:} 
    Unlearning using historical information could increase the correctness of the unlearned model, as the unlearned model is likely to converge. Nonetheless, as the model size or the number of epochs gets larger, saving all model updates is infeasible due to its high memory requirement. Thus, sampling methods should be more discovered, finding the parameters that represent and affect the model the most. 
\end{tcolorbox}

\subsubsection{Data Manipulation} 
In this approach, noise is added to the target data samples or labels to create a perturbed model \cite{xu2023revocation}.
Directly adding noise to the target samples can only be done by the target client, as the raw sample is only accessible to the data-owning client.
FedUNRAN \cite{mora2024fedunran}, for instance, replaced the true labels of the target samples with random ones.
FUCRT \cite{guo2024forgetting} changed the target samples to a transformation class, redirecting their influence within the model to semantically similar classes.
Inspired by neuroscientific principles of memory degradation, Meerza et al. \cite{meerza2024confuse} created a confusion set consisting of pairs of a target sample and a wrong label. ConFUSE thereby intentionally confused the model by adding a regularizer term to minimize predictions between the target set and the confusion set, effectively encouraging the model to forget the correct knowledge.
Conversely, Imba-ULRc \cite{yu2024federated} oversampled and generated data that remaining clients owned to weaken the influence of the target samples.

\begin{tcolorbox}[enhanced, breakable, fontupper=\linespread{.93}\selectfont]
    \textbf{\hypertarget{t7}{T-7.} Takeaways on Data Manipulation:} 
Since only the target client has access to the raw data, it can directly manipulate the target samples by adding noise or flipping labels. This confuses the global model toward mispredicting the data to be forgotten. However, this approach raises a fundamental question: does inducing incorrect predictions truly constitute unlearning?
\end{tcolorbox}

\subsubsection{Gradient Manipulation} \hfill

\textit{\underline{Perturbation:}} 
The following approaches directly introduce noise into the model parameters. This can be done both by the target client and the server, as the server also has access to the local model parameters.

In FedMUA \cite{chen2025fedmua}, the target client modified the predictions for the samples corresponding to the features that need to be forgotten.
In \cite{pan2025feature}, the server added adaptively scaled noise to the target client based on feature relevance.
FFMU \cite{che2023fast} added Gaussian noise to smooth all local models' gradients, treating them as perturbations during server aggregation.
In contrast, FedRecovery \cite{zhang2023fedrecovery} applied noise to the unlearned model to make retrained and unlearned models indistinguishable. 
SecForget \cite{liu2020learn} incorporated a trainable dummy gradient generator for each client, simulating the neurons of a model to eliminate memory of specific data.
Meanwhile, in FedFilter \cite{wang2023edge}, the server generated a random reverse gradient and performed SGD to maximize the elimination effect.

\textit{\underline{Scaling:} } 
In Verifi \cite{gao2024verifi}, target clients' gradients were downscaled, and remaining clients' gradients were upscaled, causing gradual vanishment of the target's influence. 
MoDe \cite{zhao2023federated} constructed a shadow model of the global model, sent exclusively to and trained on remaining clients. The unlearning model reduced discrimination towards target data points by adding $(1-\lambda)$ scaled weights of the shadow model to the $\lambda$, within the range [0, 1], scaled unlearning model. 
FRAMU \cite{shaik2024framu} involved clients in calculating and sending attention scores with local model updates. These scores were utilized on the server to assign less weight to model updates corresponding to the target data.
Similarly, Lin et al. \cite{lin2024blockchain} adaptively assessed the remaining clients' contributions while training the model, and the server performed a weighted average on the remaining clients' updates based on the contribution score.

\textit{\underline{Pruning:}} 
In SecureCut \cite{zhang2023securecut}, target clients pruned the nodes of their tree-structured local models before retraining them.
ElBedoui et al. \cite{elbedoui2023ecg} computed model parameters from target data and sent them to the server, which is subtracted from the global model.
Similar to \cite{xu2024update} that prune less influential channels in identifying the target class, Wang et al. \cite{wang2022federated} calculated relevant scores between channels and categories using TF-IDF and pruned the most discriminative channels of the target category. 
Similarly, FUSED \cite{Zhong2025UnlearningTK} identified the most sensitive layers through layer-wise analysis and applied an unlearning adapter that randomly dropped out some parameters within the selected layer. This approach improved efficiency by limiting unlearning and aggregation to selective layers.

\begin{tcolorbox}[enhanced, breakable, fontupper=\linespread{.93}\selectfont]
    \textbf{\hypertarget{t8}{T-8.} Takeaways on Gradient Manipulation:} 
    Models are perturbed such that they fail to achieve the task for the target information. Local updates are scaled differently to minimize and maximize the importance of the target model and retaining knowledge, respectively. Pruning refines the gradients or nodes containing the target knowledge. Nonetheless, all gradient manipulation methods incur additional complexity in finding precise perturbation, scaling factors, and pruning information because the model will not converge otherwise.
\end{tcolorbox}

\subsubsection{Loss and Influence Approximation} 
Loss functions are approximated by calculating the inverse Hessian matrix as if the unlearned model had not been trained on the target data.
The inverse Hessian, a square weighting matrix, scales gradients based on second-order partial derivatives between the training set mean point and the sample of interest \cite{mehta2022deep}. It describes the local curvature of the loss function over the entire training dataset such that the optimizer takes a more aggressive step in shallow one \cite{misra2019machine}. Nonetheless, directly calculating the inverse Hessian incurs substantial memory and computational complexity, making it infeasible in practice \cite{liu2023muter}. 

To address this, researchers have explored efficient approximation methods. Liu et al. \cite{liu2022right}, for instance, used the Quasi-Newton method, computing the Hessian matrix with the diagonal empirical Fisher Information Matrix (FIM) to estimate recovered gradients, similar to BadUnlearn \cite{wang2025poisoning} and Li et al. \cite{li2024federated}. Appro-Fun \cite{xiong2024appro} further incorporated differential privacy to mitigate data leakage.
Beyond FIM-based approximations, Xie et al. \cite{xie2024adaptive} adaptively clipped global model parameters strongly related to the target data. Meanwhile, 2F2L \cite{jin2023forgettable} relied on Taylor expansion, optimizing quadratic loss functions using neural tangent kernels and a Newton step to compute optimal weights based on the remaining dataset.

Even FedU \cite{wang2024fedu} proposed a method that estimates the influence of target data points using gradients and Hessian-Vector Products, requiring only the target samples. This method is more efficient and cost-effective since it does not depend on access to the entire remaining training dataset.

\begin{tcolorbox}[enhanced, breakable, fontupper=\linespread{.93}\selectfont]
    \textbf{\hypertarget{t9}{T-9.} Takeaways on Loss and Influence Approximation:} 
    Although an inverse Hessian allows scaling the gradients by approximating the loss function, naively calculating it incurs an expensive computational burden. Researchers thus approximate the inverse Hessian to reduce the computational complexity, although there remains more room for improvement to find a more efficient and more accurate approximation. 
\end{tcolorbox}

\subsubsection{Knowledge Distillation (KD)}
In contrast to Goldfish \cite{wang2024goldfish} and CKGD \cite{zhang2025model}, which used knowledge distillation (KD) at the server to transfer only retained knowledge to the unlearned (student) model, FedAF \cite{li2023federated} and VFU \cite{varshney2025unlearningclientsfeaturessamples} adopted a teacher-student learning pattern at the target client. In this approach, a teacher model generated fake labels, and the student model learned the manipulated knowledge. Eventually, the teacher model—never exposed to the original feature-label pair—became the unlearned model, avoiding data leakage.

FedLU \cite{zhu2023heterogeneous} introduced a loss function based on retroactive inference theory, employing mutual knowledge distillation to for unlearning while handling covariate drift between local optimization and global convergence during training. 
FedQUIT \cite{mora2024fedquit} constructed a virtual teacher to approximate the global model's knowledge without influence from the target clients, while SFU \cite{zhou2024streamlined} incorporated three teacher models: one for forgetting via negative transfer to ensure random predictions of the student model on target data and two for performance preservation. 

In a fully decentralized FU framework, HDUS \cite{ye2023heterogeneous}, all clients collaboratively generated a shared reference dataset that contained unlabeled and non-client-specific data. The remaining clients then trained their seed models on this dataset and achieved unlearning through a model ensembling. The use of seed models addressed concerns in the federated context by reducing the communication burden, minimizing data leakage risks, and allowing local models to have distinct architectures.

\begin{tcolorbox}[enhanced, breakable, fontupper=\linespread{.93}\selectfont]
    \textbf{\hypertarget{t10}{T-10.} Takeaways on knowledge distillation:} 
    Knowledge distillation reduces the training dataset or model size so that it achieves better efficiency than retraining. It is especially beneficial in that the teacher model does not have the original pair of knowledge; there would be the least residual information about the target data, providing security along with efficacy. 
\end{tcolorbox}

\subsubsection{Multi-task Learning} 
Multi-task learning balances retaining essential knowledge while discarding target knowledge in three ways. First, the unlearned models aim to maximize indistinguishability within a forget set as in Goldfish \cite{wang2024goldfish} or between it and an arbitrary dataset \cite{chen2024federated}.
For another line of work, BFU \cite{wang2023bfu} leveraged hard-parameter sharing and variational Bayesian unlearning to optimize an approximate posterior of the remaining dataset.
Similarly, Alam et al. \cite{alam2023get} focused on eliminating compromised data, augmenting loss separately for benign and spoiled data to preserve non-malicious behavior while unlearning trigger patterns. 
Finally, FedME2 \cite{xia2023fedme} optimized the local model to ensure both classification accuracy and a memory evaluation loss, derived from a lightweight evaluation model within each client, detecting whether data to be forgotten is remembered. 
FedHarmony \cite{dinsdale2022fedharmony} was trained for three tasks: to update the label predictor, to discriminate between sites, and to remove the site-specific knowledge by penalizing deviation in the probability of the outputs from a uniform distribution. 
FCU \cite{deng2024enable} employed contrastive learning; specifically, they adjusted the model parameters to minimize the difference between the current global model (negative examples) and a reference model (positive example) that has never seen the target data.

\begin{tcolorbox}[fontupper=\linespread{.93}\selectfont]
    \textbf{\hypertarget{t11}{T-11.} Takeaways on Multi-task Learning:} 
    The tasks typically involve target removal and performance restoration. It can be achieved sequentially or in parallel, where the latter achieves better time efficiency.
\end{tcolorbox}

\subsubsection{Reverse Training} 
Gradient ascent, instead of descent, is employed either stochastically by the server on the global model \cite{wu2022federated, su2024f2ul} or by the target client on the local model \cite{li2023subspace, halimi2022federated, wang2024server, ameen2024addressing, ma2024hier, han2025vertical, liu2024blockful, varshney2025unlearningclientsfeaturessamples} to reverse the original training process. This approach is different from the subtraction of historically stored parameters in that it trains a target model to achieve maximal loss through ascent.

Rather than simply flipping the gradient sign, Halimi et al. \cite{halimi2022federated} utilized projected gradient ascent to maximize local empirical loss. 
Similarly, Han et al. \cite{han2025vertical} computed ascent directions guided by the averaged gradients of the remaining clients. Building on this idea, FedOSD \cite{pan2024federated} applied gradient ascent along the orthogonal steepest descent direction to prevent conflicts with remaining clients' gradients, enhancing unlearning while preserving model utility.

In QuickDrop \cite{dhasade2024quickdrop}, each client generated a distilled dataset—a small synthetic dataset that condenses critical information from the learning phase. They subsequently fine-tuned the gradients computed from this distilled data to align with their original local datasets.


\begin{tcolorbox}[enhanced, breakable, fontupper=\linespread{.93}\selectfont]
    \textbf{\hypertarget{t12}{T-12.} Takeaways on Reverse Training:}
    Reverse training, represented by (projected/stochastic) gradient ascent, gives the most intuition for unlearning by optimizing the loss to be maximal. As it harms the original performance by a large margin, it is often accompanied by performance recovery.
\end{tcolorbox}
 
\subsubsection{Clustering} The server clustered clients to maintain the minimal number of clients to perform retraining, limited to the only cluster to which the requesting client belonged as in Liu et al. \cite{liu2024guaranteeing}.
Lin et al. \cite{lin2024scalable}, for instance, clustered the clients into multiple shards prior to training and adopted coded computation to reduce the storage overhead by compressing the model parameters across distinct shards. 
FedUMP \cite{zhu2024federated} partitioned clients based on their label distribution and unlearning cost and retrained only a subset of the partitioned clients.
Similarly, in \cite{lin2024incentive}, Hier-Fun \cite{ma2024hier}, and KNOT \cite{su2023asynchronous} grouped clients based on their similarities in model disparity (angle) and short training time to reduce the number of computations. 

\begin{tcolorbox}[fontupper=\linespread{.93}\selectfont]
    \textbf{\hypertarget{t13}{T-13.} Takeaways on Clustering:} 
This approach minimizes the number of clients requiring retraining, making it more time-efficient than involving all clients. However, finding an optimal clustering is challenging and may become infeasible with frequent unlearning requests from random clients.
\end{tcolorbox}

\subsubsection{Others}
Ferrari \cite{gu2024ferrari} minimized target feature sensitivity by quantifying the model's response to input changes using a Lipschitz continuity metric. 
Wang et al. \cite{wang2025forgettingdatatimetheoretically} proposed an unlearning method in VFL, where clients and the server maintain confidence matrices. Target clients compute differences in confidence vectors before and after unlearning, and the server updates its matrix and gradients accordingly. 
FedAU \cite{gu2024unlearning} introduced an auxiliary unlearning module from the beginning of the training process, XORing its parameters with the target model to eliminate its influence. 
TrustChain \cite{zuo2024federated} is the first, to our knowledge, to explore federated LLM unlearning. It employs blockchain to track data contributions and uses LoRA hyperparameter adjustments for targeted unlearning while ensuring transparency and accountability.

\subsection{Performance Recovery} 
Removing influence from a trained model could deteriorate its performance on the remaining dataset because target erasure often involves undoing specific parameters \cite{xu2023machine}, thereby leaving other parameters incomplete \cite{wang2022federated}. 
These performance drops become more noticeable when the model parameters are manipulated to remove influence. 
As such, additional approaches are explored to restore the performance of the unlearned model on the remaining dataset.

\subsubsection{Post Training} 
Some approaches \cite{cao2023fedrecover, gong2022forget, halimi2022federated, gong2022compressed, zhang2023securecut, ameen2024addressing, su2024f2ul, yu2024federated, mora2024fedunran, khalil2025not, han2025vertical, wang2025forgettingdatatimetheoretically} applied post-training to the unlearned model after influence removal. Wang et al. \cite{wang2024server} referred to this process as boosting training, where the remaining clients and the server perform a few additional global rounds of standard FL training--without the target influence--to restore model performance.
FedU \cite{wang2024fedu} conducted post-training simultaneously with influence removal by training the unlearned local model on the remaining local dataset.
Unlike most clustering-based unlearning techniques, Hier-Fun \cite{ma2024hier} performed a few steps of post-training with the remaining clients to restore performance, even in highly heterogeneous edge environments.
Notably, post-training does not involve additional data points.

\subsubsection{Fine-tuning} 
Building on the neurological theory that memory traces fade, some works have focused on reducing the discriminability of target data while maintaining that of the remaining data. For instance, some approaches removed the regularization term \cite{wang2022federated} or suppressed the activation of forgotten knowledge \cite{zhu2023heterogeneous}.

Zhao et al. \cite{zhao2023federated} employed guided fine-tuning, similar to MetaFul \cite{wang2023mitigating}, where the server additionally sent a degradation model to the target client. This model generated pseudo-labels, assisting the client in restoring its discriminability on the remaining data.
FUCRT \cite{guo2024forgetting} fine-tuned on a newly created dataset consisting of class-transformed target data. Specifically, the authors reassigned labels to the second most probable class, preserving the original global model’s high performance with minimal loss. During fine-tuning, they also incorporated contrastive loss to maintain feature separability.
Similarly, QuickDrop \cite{dhasade2024quickdrop} and FAST \cite{guo2023fast} used post-training on small datasets to recover performance. 
FCU \cite{deng2024enable} applied fine-tuning selectively, retaining low-frequency model components, which encode general knowledge, and adjusting high-frequency components tied to specific client data. This ensured the model could forget the targeted data without compromising overall performance.

\subsubsection{Gradient Manipulation} 
Some achieved performance recovery by directly manipulating the magnitude or direction of gradients.
Li et al. \cite{li2024federated} clipped the recovered gradients to limit errors that could degrade model performance and to preserve generalization, similar to the adaptive clipping method used by Xie et al. \cite{xie2024adaptive}.
In FRU \cite{yuan2023federated}, the remaining clients run additional local training and send the new model updates to the server. Then, the server constructed calibrated updates by combining the original local updates with the direction of the new updates. 
ConFUSE \cite{meerza2024confuse} selectively updated only salient weights during unlearning to minimize the impact on the model’s performance on retained data, achieving both performance preservation and computational efficiency.

Rather than using redirection, some methods projected the target model's gradients into the orthogonal subspace of the input space \cite{li2023subspace, pan2024federated}. If the unlearning gradients directly conflict with useful learning updates (i.e., pointing in the opposite direction), applying them without modification can disrupt the model’s overall accuracy. Projection ensures that unlearning does not interfere with the useful updates from the remaining data, minimizing performance degradation. However, precise gradient manipulation is crucial for maintaining model convergence.

\subsubsection{Regularization} 
Regularization or penalty terms, in the forms of Scaling factor \cite{liu2022right}, L1 norm \cite{liu2020learn}, or L2 norm \cite{xia2023fedme}, were applied to the unlearning model to prevent overfitting to unlearning tasks. Elastic Weight Consolidation (EWC) \cite{deng2023vertical}, FIM \cite{wu2022federated}, and the Hessian matrix \cite{li2023federated} were also often used as the regularization term to limit the magnitude of parameter updates, ensuring minimal changes to important parameters. 
Instead of a static penalty term, dynamic penalization approaches, including leveraging a momentum technique \cite{liu2022right}, were also explored \cite{alam2023get}.
Similar to \cite{deng2023vertical} that used clients' intermediate parameters computed using EWC, \cite{shao2024federated} used a penalty term derived from projecting an intermediate-term on the tangent space of the aggregated updates.

\subsubsection{Knowledge Distillation} 
Similar to influence removal, fidelity is preserved by transferring only the knowledge to be remembered to the unlearned (student) model \cite{wu2024unlearning, mora2024fedquit}. 
Wu et al. \cite{wu2024unlearning} used the original global model as a teacher to train a skewed student model, aiming to enhance both generalization and security. Additionally, the server employed unlabeled data to correct skew introduced during the influence removal process, as seen in \cite{fu2024client}, by conducting additional knowledge distillation iterations on auxiliary data at the server.

SFU \cite{zhou2024streamlined} leveraged two teacher models for performance preservation, similar to the aforementioned works, while also incorporating label-based preservation to prevent over-unlearning. The second teacher model distilled label-aware knowledge to minimize unintended information loss. Specifically, it transferred information about structural similarities in the data distribution, allowing the model to differentiate between data that should be forgotten and similar data (e.g., sharing the same label) that should be retained. As a result, the student model preserved essential decision boundaries, reducing the risk of accidental over-unlearning.

In addition to gradient clipping, Xie et al. \cite{xie2024adaptive} employed data-free knowledge distillation to restore deteriorated performance. 
They synthesized pseudo-samples to replace real data for distillation. A teacher model, trained prior to unlearning, then guided the unlearned model using soft labels on these pseudo-samples, helping recover generalization ability. This data-free knowledge distillation enables the server to guide fine-tuning without requiring access to any raw client data.


\begin{tcolorbox}[enhanced, breakable, fontupper=\linespread{.93}\selectfont]    \textbf{\hypertarget{t14}{T-14.} Takeaways on Performance Recovery:} 
    Performance recovery restores the model performance tampered with while forgetting. It is conducted sequentially or in parallel to influence removal: post-training, fine-tuning, and gradient manipulation are followed by the influence removal, while regularization using momentum or an additional term is applied during unlearning and thereafter. Note that knowledge distillation improves not only performance by generalization but also security.
\end{tcolorbox}

\section{Comparison of Evaluation Metrics}
\label{sec:evalmetric}
Unlearning methods are typically evaluated across three dimensions: efficacy (the effectiveness of unlearning/forgetting target information), fidelity (the accuracy in maintaining performance on remaining data), and efficiency (the comparison to retraining from scratch).
A variety of evaluation metrics are summarized in \autoref{tab:summary_eval_metric}. 
For brevity, we have limited the works cited here with a full reference list for each metric provided in \autoref{app:evalmetric}.

\subsection{Efficacy} 
The efficacy is defined by how accurately or effectively the unlearned model forgot the target information. 

\subsubsection{Performance Metrics}
Accuracy \cite{wang2022federated, dhasade2024quickdrop, guo2023fast, xu2023revocation, liu2021federaser, fraboni2024sifu, gao2024verifi, wu2022federated, halimi2022federated, gong2022forget, zhao2023federated, wang2023bfu, zhang2023fedrecovery, sheng2024robust, gu2024ferrari, su2024f2ul, xu2024update, ma2024hier, mora2024fedunran, zhang2025model, pan2025feature, chen2025fedmua}, 
F1 score \cite{guo2024forgetting}, loss \cite{liu2021federaser, gao2024verifi}, and error \cite{che2023fast, deng2024enable} on the forget set are commonly used to measure the efficacy of influence removal. Statistical metrics such as Mean Squared Error (MSE) and Mean Absolute Error (MAE) have also been employed \cite{shaik2024framu, liu2021revfrf}.

For recommendation tasks, in addition to Normalized Discounted Cumulative Gain at rank 10 (NDCG@10), which was used in \cite{wang2024forget}, Yuan et al. \cite{yuan2023federated} and CFRU \cite{huynh2025certified} also utilized Hit Ratio at rank 10 (HR@10) to evaluate the top-10 recommendation performance.

\subsubsection{Parameter Differences}
Kullback–Leibler divergence \cite{gao2024verifi, wang2023bfu}, L2 distance \cite{shaik2024framu, wang2023bfu}, first Wasserstein distance \cite{zhang2023securecut}, and cosine similarity \cite{nguyen2024empirical, huynh2025certified, huynh2024fast} between the unlearned and retrained models serve as indicators of unlearning efficacy. 

FedEraser \cite{liu2021federaser} measured parameter deviation $\theta$ with respect to the retrained model using: $arccos \frac{w_u w_r}{||w_u|| ||w_r||}$ where $w_u$ and $w_r$ are the last-layer weight of the unlearned and retrained model, respectively.
Symmetric Absolute Percentage Error (SAPE) \cite{liu2022right}, defined as $\frac{|Acc^u_{test} - Acc^\ast_{test}|}{|Acc^u_{test}| + |Acc^\ast_{test}|}$ was also used, where $Acc^u_{test}$ and $Acc^\ast_{test}$ denote test accuracy of retrained and unlearned model, respectively \cite{wang2024efficient, xiong2024appro}.
Gong et al. \cite{gong2022compressed} used Expected Calibration Error (ECE) to assess model calibration by quantifying the difference between actual test accuracy and confidence levels. Similarly, Elbedoui et al. \cite{elbedoui2023ecg} measured the L2 norm between confidence distributions.
Meanwhile, Shao et al. \cite{shao2024federated} designed a custom metric based on the performance difference between the unlearned and retrained models.

\subsubsection{Indiscrimination Quality}
The most commonly used metric in this category is attack-based evaluation, which assesses how effectively unlearned models mitigate Backdoor Attacks (BAs) and Membership Inference Attacks (MIAs).

In BAs, a backdoor trigger is injected into a portion of the training data, and the samples are relabeled to a predetermined class, designating them as target samples. An FL model is then trained to perform two tasks: (1) a backdoor task (mispredicting compromised samples) and (2) a main task (maintaining normal behavior on all other samples). The goal of an unlearning model is to forget backdoor-triggered data and the associated information embedded in the model. Instead of using simple BAs, Alam et al. \cite{alam2023get} employed stealthier and longer-lasting backdoors, such as Constraint-and-Scale \cite{bagdasaryan2020backdoor} and Neurotoxin \cite{zhang2022neurotoxin}.
MIAs, on the other hand, test the unlearning model's ability by identifying whether a target sample was part of the original training dataset.

Attack Success Rate (ASR) is a commonly used metric calculated as the fraction of successful attacks over total attempts. ASR represents the proportion of misclassified backdoor-triggered samples in BAs \cite{fu2024client, cao2023fedrecover, halimi2022federated, chen2025fedmua, huynh2025certified} and incorrect membership predictions in MIAs \cite{halimi2022federated, dhasade2024quickdrop, gu2024ferrari, xiong2024appro, pan2025feature}. Other evaluation metrics include precision \cite{yuan2024towards, tao2024communication, xie2024adaptive}, recall \cite{ameen2025speed, xie2024adaptive, liu2021federaser, yuan2024towards}, and F1-score \cite{lin2024scalable, lin2024incentive}.
Xu et al. \cite{xu2024update} and Ferrari \cite{gu2024ferrari} performed model inversion attacks to assess whether the forgotten target could still be reconstructed from the model parameters or gradients. Successful unlearning should drive these values toward zero.

Some works compared prediction overlap \cite{mora2024fedquit} or attention maps \cite{gu2024ferrari, zhu2024federated, zhang2025model} before and after unlearning. For instance, Ferrari \cite{gu2024ferrari} used attention maps to determine whether the unlearned model still focuses on the forgotten features. Similarly, CKGD \cite{zhang2025model} leveraged Grad-CAM to evaluate how the model's primary attention shifts after unlearning.

Meanwhile, FedHarmony \cite{dinsdale2022fedharmony} aimed to minimize domain classification accuracy while maximizing prediction accuracy. 
Verifi \cite{gao2024verifi} used influence functions to measure the residual impact of target samples on the unlearned model. 
Zhu et al. \cite{zhu2023heterogeneous} evaluated unlearning effectiveness in entity link prediction tasks by measuring Hits@N and Mean Reciprocal Rank (MRR)—where lower values on the forgetting set indicate better unlearning efficacy.

\begin{table}[tp]
\caption{A summary of Evaluation Metrics}
\label{tab:summary_eval_metric}
\setlength\tabcolsep{1pt} 
\renewcommand*{\arraystretch}{0.9}
\small
\begin{center}
\begin{tabular}{lll}
\toprule
\multicolumn{1}{c}{\textbf{Objective}} & \multicolumn{1}{l}{\textbf{Category}} & \multicolumn{1}{l}{\textbf{Metric}} \\ 
\toprule
Efficacy    & Performance   & Accuracy, F1, or AUC on the target set \\
            &               & Loss and errors on the target set  \\
            &               & MSE and MAE \\ \cmidrule{2-3}
            & Parameter     & L2 distance \\
            & difference    & KLD \\
            &               & Error rate (SAPE, ECE)  \\
            &               & Angular deviation \\
            &               & 1st Wasserstein distance \\  \cmidrule{2-3}
            & Indiscrimination  & ASR, precision, and recall on BA \\ 
            & quality           & ASR, precision, and recall on MIA \\
            &                   & Model inversion attack \\
            &                   & Multi-task learning \\
            &                   & Influence function \\ 
            &                   & Prediction overlap \\ \midrule
Fidelity    & Performance   & Accuracy, F1, or AUC  on test set  \\
            &               & Loss or errors on test set  \\ 
            &               & Accuracy, F1, or AUC  on remaining set  \\
            &               & Loss or errors on remaining set \\ \midrule
Efficiency  & Complexity    & Time taken for unlearning \\ 
            &               & Speed-up ratio  \\ 
            &               & Memory in MB \\
\bottomrule
\end{tabular}
\end{center}
\vspace{-10pt}
\end{table}

\subsection{Fidelity} 
Maintaining performance is as important as precise unlearning because otherwise, the unlearned model becomes useless. 
\textbf{Performance metrics} (i.e., accuracy, loss, and error rates) on the remaining dataset or test dataset were used to evaluate the model's fidelity. Specifically, the former is to show no drop in accuracy \cite{su2024f2ul, zhang2023fedrecovery, dhasade2024quickdrop, gu2024ferrari, pan2025feature}, or no rise in errors \cite{pan2025feature, zuo2024federated, wang2024server} on the remaining dataset, and the latter is to ensure its overall performance, including generalization capability \cite{fraboni2024sifu, varshney2024efficient, sheng2024robust, su2024f2ul, gu2024ferrari}. The higher accuracy or lower loss indicates better fidelity. 

\subsection{Efficiency} 
Computational or communications overheads are measured by comparing the time or memory taken for unlearning and retraining.
\textbf{Time taken} were measured in the number of rounds \cite{fraboni2024sifu, pan2025feature, pan2024federated, sheng2024robust}, in seconds \cite{varshney2024efficient, gu2024ferrari, zuo2025federated, xiong2024appro, ameen2025speed}, in minutes \cite{ye2023heterogeneous}, or in CPU time \cite{guo2023fast}.
A speed-up ratio is also used; similar to \cite{zhao2023federated}, \cite{cao2023fedrecover, wang2024efficient} calculated the average cost-saving percentage calculated by $\frac{T-T_r}{T}\times 100$, where $T$ denotes the number of rounds to retrain a model, and $T_r$ refers to that taken to train the unlearned model. 
\textbf{Memory} in byte \cite{liu2024guaranteeing, huynh2024fast, xu2024update, gao2024verifi} was measured to see if the unlearning method did not incur memory overhead during communication.

\tcbset{after skip balanced=0.2\baselineskip }
\begin{tcolorbox}[enhanced, breakable, fontupper=\linespread{.93}\selectfont]    \textbf{\hypertarget{t15}{T-15.} Takeaways on Evaluation Metrics.}
    Evaluations rely on indirect measurements, comparing performance, parameters, indiscrimination quality, and complexity between unlearned and retrained models. The absence of a benchmark or unified metric for "unlearnability" hampers fair comparisons across different unlearning methods using diverse architecture or aggregation methods.
\end{tcolorbox}
\begin{tcolorbox}[enhanced, breakable, fontupper=\linespread{.93}\selectfont]    \textbf{\hypertarget{t16}{T-16.} Takeaways on Attacks Used.}
    Evaluations using simple BAs are limited due to the vanishing attacks over training rounds and Byzantine-robust aggregation removing those influences. Thus, relying solely on simple BAs may obscure the impact of unlearning, making it challenging to determine its effectiveness or whether the attack has naturally diminished.
\end{tcolorbox}

\section{Gaps and Future Directions} 
\label{sec:insights}

\begin{figure*}[t]
\centerline{\includegraphics[width=\linewidth]{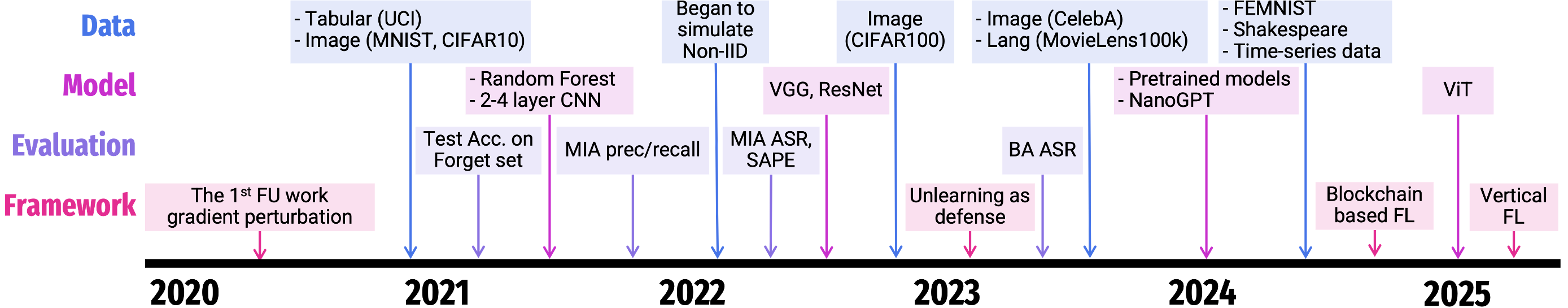}}
\caption{Emergence of the Research Focus over Time.}\label{fig:timeline} 
\vspace{-10pt}
\end{figure*}

Observing \autoref{fig:timeline}, we can see that the research focus in this area is getting diverse and deeper. Followed by the first work \cite{liu2020learn} in 2020, FU techniques have evolved from simple ML models, including Random Forest and shallow CNN. As time passes, in addition to demonstrating the effectiveness of the technique in non-IID settings, which is a common assumption in FL, more data types and, thus, more model architectures are being used, such as GPT or ViT. This year, authors are using more practical datasets, along with considering a wider range of research implications, such as fairness. 
This section summarizes research gaps and challenges that have not sufficiently been explored and potential direction.

\noindent\textbf{Lack of realistic non-IID data (\hyperlink{t1}{T-1}, \hyperlink{t2}{T-2}, \hyperlink{t3}{T-3}).}
A model successfully unlearned in an IID setting often fails to properly forget or even converge when applied to practical non-IID data, compromising both efficacy and fidelity. As in standard Federated Learning, information in FU is distributed across multiple clients, and neither the server nor other clients have access to raw data or an understanding of the distributional differences among clients.
While recent works increasingly consider heterogeneous settings, many still rely on simulated non-IID distributions generated from originally IID datasets. A commonly used simulation approach involves the Dirichlet distribution, which controls the degree of non-IIDness through a concentration parameter $\alpha$ \cite{li2022federated}. This method is widely adopted due to its ability to approximate real-world data heterogeneity. However, the choice of $\alpha$ is often arbitrary, with values ranging from 0.1 to 5 (\autoref{tab:comprehensive1}), and an optimal value that accurately reflects real-world non-IID remains unclear.

To bridge this gap, future work could explore more realistic non-IID scenarios. This includes identifying appropriate $\alpha$ values, using real-world federated datasets such as FEMNIST, Shakespeare, or StackOverflow, or constructing new benchmarks that more accurately capture the complexity of practical non-IID distributions.

\noindent\textbf{Limited application of FU methods beyond vision tasks (\hyperlink{t1}{T-1}, \hyperlink{t3}{T-3}, \hyperlink{t4}{T-4}).}
As shown in \autoref{fig:dataset} and \autoref{tab:settings}, current research on FU predominantly focuses on vision tasks—particularly image classification—while overlooking its broader applicability. This narrow focus may stem from the limited exploration of FL with LLMs, as few studies have evaluated FU methods on language tasks or datasets—even with the rise of powerful models like GPT-4. Moreover, only one study to date has utilized a multivariate dataset, highlighting a significant research gap.

Expanding FU research beyond vision tasks is both necessary and promising. Incorporating diverse tasks from other domains—such as next-word prediction for personal devices (text) or privacy-sensitive medical analysis (text, image, or multivariate data)—could showcase the true potential of FU in preserving privacy and improving efficiency over retraining.

\noindent\textbf{Difficulties in unlearning the closed-source LLMs (\hyperlink{t1}{T-1}, \hyperlink{t4}{T-4}).}
Proprietary LLMs have undergone training processes that require substantial computational resources and vast amounts of data, both public and private. To protect intellectual property and sensitive information, companies often withhold access to both the training data and model parameters—despite facing lawsuits over the use of copyrighted content in training. As data continues to evolve and removal requests grow, relying solely on post hoc fine-tuning becomes less effective due to memorization issues. Full retraining, meanwhile, is largely infeasible given the scale of modern LLMs and their datasets. 

The development trajectory of LLMs is expected to shift toward ensembling multiple pre-trained and/or new models, leveraging their powerful capabilities without exposing privacy-sensitive training data. In this context, a small, well-trained model on copyrighted data could be treated as a target client, with the original LLM acting as the global model. FU could then be employed to remove private data from the original LLM, providing a more efficient alternative to retraining and a more effective solution than fine-tuning.

\noindent\textbf{Underexplored attacks and defense methods in FU (\hyperlink{t2}{T-2}, \hyperlink{t5}{T-5}.}
The vulnerability of the FU framework has not yet been fully explored or defended against. 
Given its distributed nature, FU inherits the same risks as FL—such as poisoning and backdoor attacks—but attack and defense strategies remain under-investigated. To date, only two works \cite{sheng2024robust, wang2025poisoning} have proposed a robust FU method addressing poisoning attacks in a binary classification task.
More critically, FU may paradoxically compromise privacy despite its goal of enhancing it. Since learned and unlearned models are shared among participants, adversaries can exploit the differences to infer forgotten information through inference or model inversion attacks.

\textit{Example Attack Scenario:} Assume the remaining clients and the server are honest-but-curious. In this case, only the target client can verify if unlearning was successful. Other clients could send random or crafted parameters, which the server aggregates—potentially helping a malicious client infer the forgotten data. Worse, if the server itself is malicious, it can manipulate updates and extract sensitive data from both the target and remaining clients.

This vulnerability highlights the importance of minimizing the exact information revealed during parameter sharing to protect against inference attacks. However, no existing FU methods have been rigorously analyzed for such risks. While some works have introduced Homomorphic Encryption or differential privacy for obfuscation, establishing a trusted third party or secure execution environment could also be a viable defense strategy.

\noindent\textbf{Limited adoption of advanced FL techniques (\hyperlink{t3}{T-3}, \hyperlink{t5}{T-5}).}
Section 4 outlined the unique challenges of unlearning in the federated context. While some of these challenges—such as data heterogeneity and Byzantine failures—are actively being addressed in FL through advanced aggregation methods, many remain unexplored in FU. Notably, most existing FU works still rely on the simplistic \texttt{FedAvg}, which averages all clients’ gradients. Only a few studies—FedHarmony \cite{dinsdale2022fedharmony}, FedLU \cite{zhu2023heterogeneous}, and KNOT \cite{su2023asynchronous}—have adopted alternatives like FedProx, FedBuff, and FedEqual. Similarly, FedRecover \cite{cao2023fedrecover} and VeriFi \cite{gao2024verifi} integrated Byzantine-robust aggregation algorithms, such as Krum and trimmed-mean.

Despite these examples, only a limited number of FU methods have leveraged FL’s existing solutions, and the impact of different aggregation strategies on unlearning performance remains underexplored. It would be prudent to first assess whether established FL techniques can help mitigate FU-specific challenges before developing more complex or specialized solutions.

\noindent\textbf{Lacking consideration on scalability and fairness challenges in FU (\hyperlink{t5}{T-5}).}
In the federated context, where the number of participating clients can range from a hundred (cross-silo) to millions (cross-device), communication and computation bottlenecks can significantly impact system performance. Scalability is thus a critical consideration. Another challenge is fairness and adaptability, as clients frequently join and leave the system due to reasons like unstable network connections or attempts to exploit system instability. Even in such cases, relevant models should fully unlearn the target, requiring an approach capable of continuous and dynamic unlearning as clients and data evolve.

To mitigate malicious dropout attempts, fairness research in FL has focused on incentivizing/penalizing clients to promote voluntary participation. Similarly, FU methods need to be scalable, fair, and adaptable, ensuring effective unlearning as the client set and data evolve. This involves developing dynamic unlearning techniques that can efficiently forget specific data or clients without the need for retraining from scratch.

\noindent\textbf{Absence of a common benchmark (\hyperlink{t15}{T-15}).}
Currently, there is no standardized approach for comparing FU methods, which limits comprehensive assessment of their efficacy, fidelity, and efficiency. Most evaluations rely on indirect comparisons between retraining and unlearning performance. However, differences in model architectures, aggregation methods, and hyperparameters across studies introduce inconsistencies that hinder fair comparisons. As shown in \autoref{tab:settings}, such variations significantly influence model performance, making it difficult to reach consensus on which methods perform best.

FU research lacks benchmark datasets and unified evaluation criteria. Establishing consistent metrics and experimental setups would enable fair comparisons, helping identify the most effective methods and accelerating progress in the field.

\noindent\textbf{Unreliable assessment using naive backdoor attacks (\hyperlink{t16}{T-16}).}
Existing evaluation methods using backdoor attacks (BAs) have primarily relied on simple attacks, whose effects naturally diminish over successive training rounds. This makes it difficult to determine whether the observed mitigation is due to effective unlearning or the natural weakening of the attack. While some attacks are designed to bypass Byzantine-robust aggregation by identifying and excluding malicious gradients, only one study \cite{alam2023get} evaluated unlearning efficacy using more sophisticated BAs such as constraint-and-scale \cite{bagdasaryan2020backdoor} and Neurotoxin \cite{zhang2022neurotoxin}.

Incorporating advanced BAs is crucial for accurately demonstrating the effectiveness of unlearning, ruling out alternative explanations like natural attack decay. Future FU research should prioritize evaluating methods against a wider range of advanced BAs to provide a more rigorous and comprehensive assessment of robustness.

\noindent\textbf{Underexplored FU methods and settings}
Most existing studies assume the HFL setting, where local model parameters are exchanged. In contrast, VFL, which involves exchanging intermediate outputs between parties with feature-level partitioned data, remains largely unexplored in the context of unlearning. VFL is particularly relevant for building collaborative models among entities with aligned interests, such as banks, insurance companies, and e-commerce platforms. Despite its potential, only a few works have examined unlearning in VFL or fully decentralized FL settings.

Future research could investigate unlearning techniques applicable to different types of FL, including VFL and fully decentralized architectures. Broadening the scope to assess the efficacy and feasibility of unlearning across diverse FL scenarios could open up new possibilities for collaborative analytics without compromising data privacy. This direction holds promise for advancing FU and addressing real-world challenges in cross-organizational data collaboration.

\section{Conclusion}
While the right to be forgotten inspired MU and FU, systematic reviews of FU methods are currently limited in their number and depth. In this paper, we give special attention to the "federated" context, highlighting unique complexities, and explore the existing literature in multiple dimensions, including the entity initiating unlearning, methodologies, and limitations. We also highlight challenges in heterogeneity simulation, implications, and dataset usage statistics. This paper also provides valuable insights and directions for future research in the intricate landscape of FU. 

\bibliographystyle{IEEEtran}
\bibliography{reference}

\vfill

\newpage

\appendix{\section*{Key Terminologies Related to FL} \label{app:fl_terminology}

\begin{table*}
    \centering 
    \footnotesize
    \caption{Key Terminology Related to FL}
    \begin{tabular}{ll}
    \toprule
    \textbf{Terminology} & \textbf{Description} \\ \toprule
FL              & A decentralized ML where multiple devices train a shared model without sharing raw data. \\
Server          & A central entity orchestrates FL by distributing the global model, aggregating updates, and coordinating training. \\
Client          & Individual devices or endpoints participating in FL. Each client possesses local data to train a local model. \\
Global Model    & A model generated by aggregating clients' local model updates. \\
Local Model     & A model trained on individual clients' data, updated locally during FL rounds. \\
Aggregation     & A method to combine local model updates to create a global model \\
FedAvg          & An aggregation method averaging model parameters from participating clients. \\
Asynchronous FL & Allows clients to update the global model without synchronization at different timings. \\
Cross-silo FL   & A type of FL that trains a global model across 2-100 data silos. \\
Cross-device FL & A type of FL that trains a global model up to $10^{10}$ clients (mobile or IoT devices). \\
VFL             & A type of FL that trains a global model across vertically partitioned data sources with different features but shared sample IDs. \\
HFL             & A type of FL that trains a global model across horizontally partitioned data sources with the same features but different samples. \\
IID data        & Data sampled from the identical and independent distribution. \\
Non-IID data    & Data sampled from not identical or not independent distribution across clients. Also known as heterogeneous data. \\
HE              & Homomorphic Encryption. A cryptographic technique that allows computation on the encrypted model parameters. \\
DP              & Differential Privacy. Ensures the presence or absence of an individual's data does not significantly affect the output. \\
    \bottomrule
    \end{tabular}
    \label{tab:terminology}
\end{table*}

\autoref{tab:terminology} summarizes the key terminologies and descriptions, including FL components, various types of FL, data distribution, and standard privacy-preserving techniques.

\section*{Evaluation Metric} \label{app:evalmetric}
In ~\autoref{tab:eval_metric_w_ref}, we included the reference for each metric. The predominant metric used for evaluating efficacy was accuracy on the target dataset, followed by the attack success rate on backdoor attacks. For assessing fidelity, the majority of works measured accuracy on the test dataset, while efficiency was predominantly evaluated based on time reduction in seconds or the number of rounds.

\begin{table*}[htbp]
\caption{A summary of Evaluation Metrics with References}
\label{tab:eval_metric_w_ref}
\footnotesize
\begin{center}
\begin{tabular}{L{1.5cm}L{2cm}L{4cm}L{9cm}}
\toprule
\textbf{Objective} & \textbf{Category} & \textbf{Metric} & \textbf{Reference} \\ \toprule
Efficacy    & Performance   & Accuracy on the target set & \cite{chen2024federated, wang2022federated, dhasade2024quickdrop, guo2023fast, xu2023revocation, liu2021federaser, fraboni2024sifu, gao2024verifi, wu2022federated, halimi2022federated, jin2023forgettable, gong2022forget, zhao2023federated, wang2023bfu, zhang2023fedrecovery, sheng2024robust, chundawat2024conda, gu2024ferrari, su2024f2ul, xu2024update, zhou2024streamlined, gu2024unlearning, zuo2024federated, ma2024hier, yu2024federated, guo2024forgetting, liu2024blockful, mora2024fedunran, zuo2025federated, mora2024fedquit, khalil2025not, zhang2025model, Zhong2025UnlearningTK, pan2025feature, chen2025fedmua}\\
            &               & F1 or AUC on the target set & \cite{guo2024forgetting} \\
            &               & Loss or errors on the target set & \cite{liu2021federaser, deng2024enable, gao2024verifi, che2023fast}  \\
            &               & MSE and MAE & \cite{shaik2024framu, liu2021revfrf}\\ \cmidrule{2-4}
            & Difference    & L2 distance & \cite{shaik2024framu, wang2023bfu, elbedoui2023ecg} \\
            &               & KLD & \cite{gao2024verifi, wang2023bfu} \\
            &               & Error rate (SAPE, ECE) & \cite{liu2022right, gong2022compressed, wang2024efficient, xiong2024appro}  \\
            &               & Angular deviation  & \cite{cao2023fedrecover, lin2024blockchain, nguyen2024empirical, huynh2025certified, huynh2024fast} \\
            &               & Wasserstein distance & \cite{zhang2023securecut} \\ \cmidrule{2-4}
            & Indiscrimination quality & ASR, precision, or recall on BA & \cite{nguyen2024empirical, fu2024client, liu2024privacy, wang2024goldfish, cao2023fedrecover, li2023subspace, wu2024unlearning, jin2023forgettable, zhao2023federated, li2023federated, che2023fast, halimi2022federated, alam2023get, wang2023bfu, deng2023vertical, jiang2024towards, sheng2024robust, jiang2024efficient, chundawat2024conda, huynh2024fast, zhou2024streamlined, gu2024unlearning, pan2024federated, wang2024fedu, meerza2024confuse, li2024federated, jiang2024feduhb, han2025vertical, wang2025poisoning, chen2025fedmua, huynh2025certified} \\ 
            
            &                  & ASR, precision, or recall on MIA & \cite{lin2024incentive, yuan2024towards, zhang2023fedrecovery, wang2022federated, halimi2022federated, fu2024client, chen2024federated, liu2024privacy, dhasade2024quickdrop, tao2024communication, jiang2024towards, lin2024blockchain, liu2021federaser, lin2024scalable, jiang2024efficient, sheng2024robust, chundawat2024conda, gu2024ferrari, xu2024update, gu2024unlearning, meerza2024confuse, guo2024forgetting, xie2024adaptive, mora2024fedunran, jiang2024feduhb, mora2024fedquit, xiong2024appro, khalil2025not, han2025vertical, ameen2025speed, varshney2025unlearningclientsfeaturessamples, Zhong2025UnlearningTK, pan2025feature} \\
            &                   & Model Inversion attack & \cite{xu2024update, gu2024ferrari} \\
            &                   & Multi-task learning & \cite{dinsdale2022fedharmony} \\
            &                   & Influence function & \cite{gao2024verifi} \\ 
            &                   & Prediction overlap & \cite{mora2024fedquit}, Grad-CAM \cite{zhang2025model}, attention map \cite{gu2024ferrari}, activation map \cite{zhu2024federated} \\  \midrule
Fidelity    & Performance   & Accuracy on test set  & \cite{liu2021federaser, jin2023forgettable, lin2024scalable, lin2024incentive, lin2024blockchain, li2023federated, deng2023vertical, gong2022compressed, fraboni2024sifu, che2023fast, elbedoui2023ecg, jiang2024towards, tao2024communication, chen2024federated, wang2023edge, fu2024client, cao2023fedrecover, su2023asynchronous, halimi2022federated, wang2023mitigating, wang2024goldfish, wang2024efficient, wu2024unlearning, varshney2024efficient, jiang2024efficient, sheng2024robust, ye2023heterogeneous, yuan2024towards, zhao2023federated, zhang2023securecut, wu2022federated, ameen2024addressing, su2024f2ul, gu2024ferrari, pan2024federated, li2024federated, zuo2024federated, guo2024forgetting, zhu2024federated, xie2024adaptive, yu2024federated, wang2024fedu, deng2024enable, mora2024fedunran, mora2024fedquit, khalil2025not, ameen2025speed, wang2025forgettingdatatimetheoretically} \\
            &               & F1 or AUC on test set & \cite{wang2025forgettingdatatimetheoretically} \\ 

            &               & Loss or errors on test set & \cite{ameen2024addressing, jiang2024feduhb, xiong2024appro, wang2025poisoning, varshney2025unlearningclientsfeaturessamples, liu2021federaser, guo2023fast, wang2023mitigating, lin2024blockchain, lin2024scalable, ameen2025speed} \\

            &               & Accuracy on remaining dataset  & \cite{su2024f2ul, wang2022federated, wu2022federated, jin2023forgettable, chen2024federated, li2023subspace, zhao2023federated, xia2023fedme, alam2023get, zhang2023fedrecovery, dhasade2024quickdrop, xu2023revocation, chundawat2024conda, wang2024server, liu2024guaranteeing, huynh2024fast, gu2024ferrari, pan2025feature, xu2024update, deng2024enable, zhou2024streamlined, gu2024unlearning, wang2024fedu, meerza2024confuse, ma2024hier, guo2024forgetting, Zhong2025UnlearningTK, chen2025fedmua, liu2024blockful, mora2024fedquit, khalil2025not, han2025vertical, zhang2025model} \\ 
            &               & F1 or AUC on remaining set & \cite{liu2021revfrf, deng2024enable, varshney2025unlearningclientsfeaturessamples, pan2025feature, guo2024forgetting} \\
            &               & Loss or errors on remaining set & \cite{pan2025feature, zuo2024federated, che2023fast, wang2024server, zhang2025model,  deng2024enable} \\  \midrule
Efficiency  & Complexity    & Time taken (in \# rounds, s, min) & \cite{fraboni2024sifu, tao2024communication, jiang2024towards, gao2024verifi, su2023asynchronous, li2023federated, zhang2023fedrecovery, che2023fast, fu2024client, wang2024efficient, yuan2024towards, chen2024federated, liu2024privacy, liu2021federaser, liu2022right, halimi2022federated, dhasade2024quickdrop, guo2023fast, lin2024blockchain, lin2024scalable, ye2023heterogeneous, sheng2024robust, varshney2024efficient, chundawat2024conda, wu2024unlearning, ameen2024addressing, wang2024server, liu2024guaranteeing, huynh2024fast, gu2024ferrari, xu2024update, deng2024enable, zhou2024streamlined, gu2024unlearning, pan2024federated, wang2024fedu, zuo2024federated, ma2024hier, zhu2024federated, liu2024blockful, mora2024fedunran, xie2024adaptive, zuo2025federated, jiang2024feduhb, xiong2024appro, han2025vertical, ameen2025speed, Zhong2025UnlearningTK, wang2025forgettingdatatimetheoretically, huynh2025certified} \\
            &               & Speed-up ratio & \cite{cao2023fedrecover, zhao2023federated, guo2024forgetting} \\ 
            &               & Memory in MB/GB & \cite{liu2021revfrf, guo2023fast, lin2024scalable, liu2024guaranteeing, wang2025forgettingdatatimetheoretically, Zhong2025UnlearningTK, xie2024adaptive, huynh2024fast, xu2024update, gao2024verifi, liu2024privacy, huynh2025certified}\\
\bottomrule
\end{tabular}
\end{center}
\vspace{-10pt}
\end{table*}

\section*{Comprehensive Tables} \label{app:comprehensive}

Across ~\autoref{tab:comprehensive1} to ~\autoref{tab:comprehensive3}, we summarized all the published literature on FU in multiple dimensions of assumptions, unlearning methods, their configurations, and links to the code, if released. It would give the readers a more comprehensive view of all the works and make it easy to compare them. 

\makeatletter \@rot@twosidefalse \makeatother
\begin{sidewaystable*}
    \caption{A comprehensive table summarizing the literature on FU. "Remaining Acc." and " Target Acc." are short for accuracy on remaining and target data, respectively. "n/d" represents not disclosed within the paper, and "-" indicates none. The numbers in brackets in the "Data Dist." column indicate the range of the concentration parameter they used. Evaluation metrics for each objective are separated by ";." Note that we added links to the code if they are released.}
    \label{tab:comprehensive1}
    \footnotesize
    \setlength\tabcolsep{2.2pt} 
    \begin{tabular}{C{0.3cm}L{1.5cm}L{1cm} 
    L{3.8cm}L{3.75cm} 
    L{2.8cm}L{2.3cm} 
    L{2cm}L{1.6cm} 
    C{1.5cm}C{1.5cm}C{0.5cm}} 
    \toprule
    \multicolumn{1}{c}{\multirow{2}*{\textbf{Ref.}}} & \multicolumn{1}{c}{\multirow{2}*{\textbf{Unlearner}}} & \multicolumn{1}{c}{\multirow{2}*{\textbf{Target}}}
    & \multicolumn{1}{c}{\multirow{2}*{\textbf{Influence removal}}} & \multicolumn{1}{c}{\multirow{2}*{\textbf{Performance recovery}}} 
    & \multicolumn{1}{c}{\multirow{2}*{\textbf{Implication}}} & \multicolumn{1}{c}{\multirow{2}*{\textbf{Eval. Metric}}}
    & \multicolumn{1}{c}{\multirow{2}*{\textbf{Dataset}}} & \multicolumn{1}{c}{\multirow{2}*{\textbf{Data Dist.}}} 
    & \textbf{Agg. Method} & \textbf{Model Arch.} & \multicolumn{1}{c}{\multirow{2}*{\textbf{Code}}} 
    \\ \toprule
    
\cite{liu2021revfrf} & Server & Client 
& Remove the node of the target client and all the child nodes 
& - (removing only a few (<=5) clients does not influence the efficacy of RevFRF) 
& Efficacy, Fidelity, Security, Guarantee 
& MSE, MAE, R-Square; Remaining Acc., recall, F1; time(s), memory(MB) 
& UCI tabular datasets \footnotemark{}  & n/d 
& n/d & Random forest & - \\ \midrule

\cite{xiong2023exact} & All clients & Sample 
& Retrain the remaining clients (exact unlearning) using a quantized learning model 
& - 
& Efficacy, Fidelity, Efficiency 
& ASR (MIA), SAPE on the unlearned model; Speed up ratio 
& fMNIST, CIFAR10 & Non-IID (random) 
& FedAvg & 3, 4-layer CNN & \href{http://www.dropbox.com/sh/pdgl4vfbbhdxxml/AADBJeS6JKCAfw5TCD4Oe1Oya?dl=0}{[link]} \\ \midrule


\cite{wang2024efficient} & Server & Feature
& Retrain from historically saved local model checkpoints
& Automatic optimizer control for generalization
& Efficacy, Efficiency, Guarantee, Adaptivity
& SAPE; Test Acc.; Speed-up ratio
& MNIST, CIFAR10, Speechcommand, ImageNet, BrainTumorMRI, ModelNet & n/d
& n/d & LeNet5, ResNet18, M5, ResNet34, pointConv & - \\ \midrule


\cite{wang2024forget} & Remaining clients & Client 
& Calibrate the local update using the stored historical local updates' direction
& -
& Efficacy, Fidelity
& NDCG@10
& MQ2007, MSLR-WEB10k, Yahoo, Istella-S & IID
& FedAvg & Linear model & \href{https://github.com/ielab/2024-ECIR-foltr-unlearning}{[link]} \\ \midrule

\cite{cao2023fedrecover} & Server, Remaining clients & Client 
& Recover from historically saved remaining local model updates
& Pre- and Post-training: the remaining clients train and send the updates in the first and last several rounds 
& Efficacy, Efficiency, Security 
& ASR (BA); Test error; time saved \% (\# rounds)
& MNIST, fMNIST, Purchase, HAR & Non-IID (Fang [0.1,1]) 
& FedAvg, median, trimmed-mean & 3-layer CNN, FCNN & - \\ \midrule

\cite{wu2024unlearning} & Server & Sample 
& Recover from historically saved remaining local model updates by averaging them and subtracting the target clients' update
& KD: using the old global model as a teacher model, transfer remembered knowledge to the skewed unlearning model (student model) 
& Efficacy, Fidelity 
& ASR (BA); Test Acc
& MNIST, CIFAR10, GTSRB & n/d 
& FedAvg & 2-layer CNN, VGG11, AlexNet & - \\ \midrule

\cite{zhang2023fedrecovery} & Server & Sample 
& Recover from historically saved local model updates by removing a weighted sum of them from the global model
& Add Gaussian noise for indistinguishability of the retrained and unlearned model 
& Efficacy, Fidelity, Efficiency, Security, Guarantee 
& Target Acc., ASR (MIA); Remaining Acc.; time (s)
& MNIST, CIFAR10, SVHN, USPS & IID 
& FedAvg & pre-trained CNN & - \\ \midrule

\cite{deng2023vertical} & Server & Client, Feature 
& Recover from saved target model updates at the last round by subtracting them from the global model
& Constraints on intermediate local model parameter during unlearning
& Efficacy, Fidelity, Efficiency, Security 
& ASR (data poisoning); acc on test dataset; time (\# rounds)
& Cood-RNA, Iris, Adult, Breast-cancer & IID 
& n/d & CNN & - \\ \midrule


\cite{wang2023mitigating} & Target Client & Sample, Client 
& Recover from historically saved updates and subtract the target model updates
& Fine-tune the unlearned model using the direction of the target updates  
& Efficacy, Fidelity, Efficiency, Adaptivity
& MAPE, Test Acc. and Loss 
& Pix3D, VRHP & IID, Non-IID (Dirichlet 0.1)
& FedAvg & VGG16, LSTM & -\\ \midrule

\cite{liu2021federaser} & Remaining clients & Client 
& Recover from periodically saved local model updates by applying weighted averages to them
& Post-training: a few training rounds after unlearning to redirect the updates without the target client 
& Efficacy, Fidelity, Efficiency 
& Prediction difference, attack precision and recall (MIA), Target Acc., loss; Remaining Acc., loss; time(s)
& Adult, Purchase, MNIST, CIFAR10 & n/d & FedAvg & 2-, 3-layer NN, 4-layer CNN & \href{http://www.dropbox.com/s/1lhx962axovbbom/FedEraser-Code.zip?dl=0}{[link]} \\ 

    \bottomrule
    \end{tabular}
\end{sidewaystable*}
\footnotetext{Adult, bank market, drug consumption, wine quality (classification), superconduct, appliance energy, insurance company, news popularity (regression)}

\makeatletter \@rot@twosidefalse \makeatother
\begin{sidewaystable*}
    \caption{A comprehensive table summarizing the literature on FU. (continued)}
    \label{tab:comprehensive2}
    \footnotesize
    \setlength\tabcolsep{2.2pt} 
    \begin{tabular}{C{0.3cm}L{1.5cm}L{1cm} 
    L{3.8cm}L{3.75cm} 
    L{2.7cm}L{2.4cm} 
    L{2.2cm}L{1.6cm} 
    C{1.5cm}C{1.5cm}C{0.5cm}} 
    \toprule
    \multicolumn{1}{c}{\multirow{2}*{\textbf{Ref.}}} & \multicolumn{1}{c}{\multirow{2}*{\textbf{Unlearner}}} & \multicolumn{1}{c}{\multirow{2}*{\textbf{Target}}}
    & \multicolumn{1}{c}{\multirow{2}*{\textbf{Influence removal}}} & \multicolumn{1}{c}{\multirow{2}*{\textbf{Performance recovery}}} 
    & \multicolumn{1}{c}{\multirow{2}*{\textbf{Implication}}} & \multicolumn{1}{c}{\multirow{2}*{\textbf{Eval. Metric}}}
    & \multicolumn{1}{c}{\multirow{2}*{\textbf{Dataset}}} & \multicolumn{1}{c}{\multirow{2}*{\textbf{Data Dist.}}} 
    & \textbf{Agg. Method} & \textbf{Model Arch.} & \multicolumn{1}{c}{\multirow{2}*{\textbf{Code}}} 
    \\ \toprule


\cite{yuan2023federated} & Server, All clients & Client 
& Restore certain global model updates to roll back to the point before the target client joined
& Gradient manipulation: calibrate the unlearning model to the historical model updates’ direction 
& Efficacy, Fidelity, Efficiency, Scalability 
& HR@10, NDCG@10 
& MovieLens-100k, Steam-200k & n/d & FedAvg & NCF, LightGCN & -\\ \midrule

\cite{fraboni2024sifu} & Server, Remaining clients & Client 
& Restore and perturb the global model of an optimal iteration
& - 
& Efficacy, Fidelity, Efficiency, Security, Guarantee
& Target Acc.; time (\# rounds)
& MNIST, CIFAR, CelebA & IID, Non-IID (Dirichlet $\alpha=1$) 
& FedAvg & Logistic regression model, CNN & \href{http://github.com/Accenture/Labs-Federated-Learning/tree/SIFU}{[link]} \\ \midrule

\cite{yuan2024towards} & Server & Client
& Adjust direction from historically saved local updates only if the model performance drops
& -
& Efficacy, Fidelity, Efficiency, Adaptivity
& Prec., Recall (MIA); Test Acc., Time (s)
& MNIST, CIFAR10, SVHN & n/d
& n/d & n/d & - \\ \midrule

\cite{nguyen2024empirical} & Server & Client
& Perturb target label to the fake one
& -
& Efficacy, Fidelity
& ASR (BA), Cosine simm.
& MNIST, CIFAR10, CIFAR100 & IID
& FedAvg & 3-layer CNN, ResNet18
& \href{https://github.com/sail-research/fed-unlearn}{[link]} \\ \midrule 
    

\cite{che2023fast} & Target client & Sample 
& Perturbation (Gaussian noise) to the target client
& - 
& Efficacy, Fidelity, Efficiency, Guarantee
& Errors on target data; errors on remaining/test data, Test Acc.; time (s)
& fMNIST, CIFAR10, SVHN & n/d 
& FedAvg & CNN, LeNet, ResNet18 & - \\ \midrule

\cite{wang2023edge} & Server & Sample
& Perturbing target update by generating a reverse gradient 
& SGD to minimize the impact on the model accuracy 
& Efficacy
& Test Acc. 
& MovieLens-100k & Non-IID
& Avg. base layers & 4-layer CNN & -\\ \midrule

\cite{xu2023revocation} & Target client & Client
& Perturbing target updates by training it on noised input
& - 
& Efficacy, Efficiency, Security
& Target Acc.; Remaining Acc. 
& MNIST & Non-IID (random)
& FedAvg & DNN & - \\ \midrule

\cite{zhao2023federated} & Server, target client & Class, Client 
& Scale down on the target data points gradually using the momentum technique
& Guided fine-tuning on remaining data points' direction
& Efficacy, Fidelity, Efficiency 
& Target Acc., ASR (BA); Remaining Acc.; speed up ratio 
& MNIST, fMNIST, CIFAR10, CIFAR100 & Non-IID (Dirichlet) 
& FedAvg & ResNet18, ResNet50 & - \\ \midrule

\cite{shaik2024framu} & All clients & Sample
& Scale up/down using an attention mechanism to assign reduced weights to the target data
& - 
& Efficacy, Fidelity, Adaptivity 
& MSE, MAE, L2 distance 
& AMPds2, METR-LA, MIMIC-III, NYPD, MIMIC-CXR, SHED & Non-IID (concept drift)
& FedAvg & n/d & - \\ \midrule

\cite{gao2024verifi} & Server & Client 
& Scale up/down on remaining/target clients' updates
& - 
& Efficacy, Fidelity, Efficiency, Security 
& Target Acc. and loss, IF, KLD; time (s), space (MB) 
& MNIST, CIFAR10, SpeechCommand, ISIC, COVID, ImageNet, VGGFace\footnotemark[9]{} & IID, Non-IID (Dirichlet [0.5, 0.9]
& FedAvg, Krum, Median & LeNet5, ResNet18, CNN-LSTM, DenseNet121 & - \\ \midrule


 
\cite{wang2022federated} & Server and all clients & Class
& Prune the channel of gradients corresponding to the target 
& Fine-tuning: few FL training with remaining clients without regularization 
& Efficacy, Fidelity, Efficiency 
& Target Acc., ASR (MIA); Remaining Acc.; speed up ratio 
& CIFAR10, CIFAR100 & IID, Non-IID (Fang) 
& FedAvg & ResNet, pre-trained VGG & \href{http://github.com/IMoonKeyBoy/Federated-Unlearning-via-Class-Discriminative-Pruning}{[link]} \\ \midrule

\cite{zhang2023securecut} & Target clients & Sample, Feature 
& Prune the nodes of the target's local model and retrain it
& Post-training: a few training rounds after unlearning 
& Efficacy, Fidelity, Security 
& 1st Wasserstein distance; Test Acc.
& Give credit, optDigits, Epsilon & n/d 
& n/d & GBDT & - \\ \midrule

\cite{guo2023fast} & Server & Client
& Subtract target updates from the global model
& Fine-tuning with a small (1/8) sized IID benchmark dataset 
& Efficacy, Fidelity, Efficiency, Security
& Target Acc., F1 score; Test Acc., loss; time(CPU, s), memory(MB)
& MNIST, fMNIST, CIFAR10, SVHN & IID, Non-IID (random)
& FedAvg & MLP, 2-layer CNN, VGG11, MobileNet & - \\  \midrule

\cite{elbedoui2023ecg} & Target client & Sample, Client
& Subtract target model updates computed only using the target data from the global model
& - 
& Efficacy
& L2 distance in confidence; Test Acc., loss
& MIT-BIH, BIDMC & IID
& FedAvg & 3-layer CNN & - \\ \midrule

\cite{jin2023forgettable} & Target client & Client 
& Loss function approximation using Hessian matrix calculated on data the server owns
& -
& Efficacy, Fidelity 
& Target Acc., ASR (MIA, BA); Remaining Acc., Test Acc.
& MNIST, fMNIST & IID 
& FedAvg & 3-layer CNN & - \\ 

    \bottomrule
    \end{tabular}
\end{sidewaystable*}

\makeatletter \@rot@twosidefalse \makeatother
\begin{sidewaystable*}
    \caption{A comprehensive table summarizing the literature on FU. (continued)}
    \label{tab:comprehensive3}
    \footnotesize
    \setlength\tabcolsep{2.2pt} 
    \begin{tabular}{C{0.3cm}L{1.5cm}L{1cm} 
    L{4cm}L{3.9cm} 
    L{2.5cm}L{2.4cm} 
    L{2.1cm}L{1.5cm} 
    C{1.3cm}C{1.7cm}C{0.5cm}} 
    \toprule
    \multicolumn{1}{c}{\multirow{2}*{\textbf{Ref.}}} & \multicolumn{1}{c}{\multirow{2}*{\textbf{Unlearner}}} & \multicolumn{1}{c}{\multirow{2}*{\textbf{Target}}}
    & \multicolumn{1}{c}{\multirow{2}*{\textbf{Influence removal}}} & \multicolumn{1}{c}{\multirow{2}*{\textbf{Performance recovery}}} 
    & \multicolumn{1}{c}{\multirow{2}*{\textbf{Implication}}} & \multicolumn{1}{c}{\multirow{2}*{\textbf{Eval. Metric}}}
    & \multicolumn{1}{c}{\multirow{2}*{\textbf{Dataset}}} & \multicolumn{1}{c}{\multirow{2}*{\textbf{Data Dist.}}} 
    & \textbf{Agg. Method} & \textbf{Model Arch.} & \multicolumn{1}{c}{\multirow{2}*{\textbf{Code}}} 
    \\ \toprule

\cite{liu2022right} & Target client & Client 
& Loss function approximation relying on Hessian using diagonal FIM
& Apply momentum techniques to the Hessian Diagonal
& Efficacy, Fidelity, Efficiency, Guarantee, Flexibility
& SAPE; Remaining Acc.; time (s)
& MNIST, fMNIST, CIFAR10, CelebA & IID 
& FedAvg & 3-layer CNN, AlexNet, ResNet & \href{http://github.com/IMoonKeyBoy/The-Right-to-be-Forgotten-in-Federated-Learning-An-Efficient-Realization-with-Rapid-Retraining}{[link]}  \\ \midrule

\cite{xia2023fedme} & All clients & Feature 
& Multi-task learning: data erasure and remembrance by optimizing local model loss using MIA-like evaluation model
& Add a regularization term: L2 norm between the local and global models  
& Efficacy, Efficiency, Guarantee
& Target Acc.; Remaining Acc.
& SVHN, CelebA & n/d 
& FedAvg & MobileNetV3, ResNet, RegNet\footnotemark[11]& - \\ \midrule


\cite{wang2023bfu} & Target client & Sample 
& Multi-task: The target client performs multi-task learning (data erasure). 
& Multi-task: The target client performs multi-task learning (maintaining acc on the remaining set) 
& Efficacy, Fidelity, Efficiency 
& ASR (BA), L2 norm distance, KLD; Test Acc.; time (s)
& MNIST, CIFAR10 & n/d 
& FedAvg & 3-layer BNN, ResNet18 & \href{http://github.com/wwq5-code/BFU-Code}{[link]} \\ \midrule

\cite{dinsdale2022fedharmony} & All clients & Feature 
& Multi-task learning: data erasure using losses to discriminate between the sites and to remove site-specific information
& Multi-task learning: remembrance using main task loss for classification
& Efficacy 
& MAE, SCA
& ABIDE & Non-IID (covariate shift) 
& FedEqual & VGG-based CNN & \href{http://github.com/nkdinsdale/FedHarmony}{[link]}  \\ \midrule

\cite{dhasade2024quickdrop} & All clients & Class, Client 
& Reverse training (stochastic gradient ascent) on the distilled dataset that each client has created
& Fine-tuning the unlearned model by including a few original data samples into the training dataset
& Efficacy, Fidelity, Efficiency, Scalability 
& Target Acc., Target ASR (MIA); Remaining ASR (MIA); time (s)
& MNIST, CIFAR10, SVHN & IID, Non-IID (Dirichlet $\alpha=0.1$) 
& FedAvg & 3-layer ConvNet & - \\ \midrule

\cite{gong2022forget} & Target client & Sample 
& Variational inference, maximizing local loss 
& Post-training: a few training rounds after unlearning 
& Efficacy, Fidelity 
& variational posterior, Target Acc. 
& MNIST & Non-IID (unique) 
& n/d & 1-layer BNN & - \\ \midrule

\cite{gong2022compressed} & Target client & Sample, Class, Client 
& Variational inference, quantization and sparsification method to reduce communication overhead
& Post-training: a few training rounds after unlearning with remaining clients
& Efficacy, Fidelity, Efficiency 
& Target Acc., ECE; Remaining Acc. 
& fMNIST & Non-IID (unique) 
& FedAvg & NN & - \\ \midrule

\cite{su2023asynchronous} & Server, Target client, Some remaining clients & Client 
& Cluster clients based on their similarity (training time and low model disparity) and retrain the only cluster that contains the target client. 
& - 
& Efficacy, Fidelity Efficiency, Scalability 
& Test Acc.; time (s)
& CIFAR10, EMNIST, Purchase 100, Tiny Shakespeare & Non-IID (Dirichlet $\alpha=5$) 
& FedAvg, FedBuff & VGG16, LeNet5, MLP, GPT2 & \href{http://github.com/TL-System/plato/tree/main/examples/knot}{[link]} \\ \midrule

\cite{lin2024incentive} & Server & Client
& Cluster based on their similarity and calibrate the gradient by step length and direction
& Dynamically select remaining clients possessing unbalanced local data
& Efficacy, Fidelity, Fairness
& F1 (MIA); Test Acc. 
& MNIST, fMNIST, CIFAR10 & IID, Non-IID
& FedAvg & n/d & - \\ \midrule

\cite{zhu2023heterogeneous} & Server, All clients & Sample 
& Knowledge distillation: transfer knowledge to be remembered
& Fine-tuning: passive decay by suppressing the activation of the target data
& Efficacy, Fidelity, Efficiency 
& Average Hits@N, MRR 
& FB15k-237 & Non-IID (unique)
& FedAvg & TransE, ComplEx, RotE & \href{https://github.com/nju-websoft/FedLU/}{[link]} \\ \midrule

\cite{li2023federated} & Target clients & Sample, Class, Client 
& Knowledge distillation: the target client’s student model distills knowledge from manipulated data (fake labels with original features) that teacher models generated.
& EWC training, using a Hessian matrix as a regularizer
& Efficacy, Fidelity, Efficiency 
& ASR (BA); Test Acc.; time(s)
& MNIST, CIFAR10, CelebA & n/d 
& FedAvg & 3-layer CNN, ResNet10 & - \\ \midrule

\cite{ye2023heterogeneous} & Remaining clients & Class 
& Ensemble remaining clients' knowledge-distilled reference models
& - 
& Efficacy, Scalability, Flexibility
& Test Acc., time (min) 
& MNIST, FMNIST, CIFAR10 & Non-IID (unique) 
& n/d & ResNet, MobileNet & - \\ \midrule

\cite{wu2022federated} & Target client & Sample, Class, Client 
& Reverse training: stochastic gradient ascent 
& EWC training, limit the magnitude of parameters through FIM as a regularizer 
& Efficacy, Fidelity 
& Target Acc.; Remaining and Test Acc.
& MNIST, CIFAR10 & IID, Non-IID (unique)
& FedAvg & n/d & - \\ \midrule

\cite{li2023subspace} & Server, All clients & Sample 
& Reverse training: Gradient Ascent (GA)
& Server projecting target updates into the orthogonal subspace calculated from remaining clients' representation matrices
& Efficacy, Fidelity, Security
& ASR (BA); Test Acc.
& MNIST, CIFAR10, CIFAR100 & IID, Non-IID (Dirichlet [0.1, 0.6])
& n/d & MLP, 3-layer CNN, ResNet18 & - \\ 

\bottomrule
    \end{tabular}
\end{sidewaystable*}
\footnotetext[15]{ImageNet and VGGFace are ImageNet\_mini, VGGFace\_mini}

\makeatletter \@rot@twosidefalse \makeatother
\begin{sidewaystable*}
    \caption{A comprehensive table summarizing the literature on FU. (continued)}
    \label{tab:comprehensive3}
    \footnotesize
    \setlength\tabcolsep{2.2pt} 
    \begin{tabular}{C{0.3cm}L{1.5cm}L{1cm} 
    L{4cm}L{3.9cm} 
    L{2.5cm}L{2.4cm} 
    L{2.1cm}L{1.5cm} 
    C{1.3cm}C{1.7cm}C{0.5cm}} 
    \toprule
    \multicolumn{1}{c}{\multirow{2}*{\textbf{Ref.}}} & \multicolumn{1}{c}{\multirow{2}*{\textbf{Unlearner}}} & \multicolumn{1}{c}{\multirow{2}*{\textbf{Target}}}
    & \multicolumn{1}{c}{\multirow{2}*{\textbf{Influence removal}}} & \multicolumn{1}{c}{\multirow{2}*{\textbf{Performance recovery}}} 
    & \multicolumn{1}{c}{\multirow{2}*{\textbf{Implication}}} & \multicolumn{1}{c}{\multirow{2}*{\textbf{Eval. Metric}}}
    & \multicolumn{1}{c}{\multirow{2}*{\textbf{Dataset}}} & \multicolumn{1}{c}{\multirow{2}*{\textbf{Data Dist.}}} 
    & \textbf{Agg. Method} & \textbf{Model Arch.} & \multicolumn{1}{c}{\multirow{2}*{\textbf{Code}}} 
    \\ \toprule 
    
\cite{wu2024unlearning} & Server & Client
& Recover from historically saved remaining client updates
& KD: transfer remembered knowledge
& Efficacy, Fidelity, Efficacy
& ASR (BA); Test Acc.; Time (s)
& MNIST, CIFAR10, CTSRB, IMDb & IID, Non-IID (Dirichlet [0.1, 1])
& FedAvg & 3-layer CNN, VGG9, 16, ResNet, MobileNetv2, AlexNet & - \\ \midrule

\cite{ameen2024addressing} & Server, Target client & Sample
& Reverse Training: Layer-wise GA 
& Post-training
& Efficacy, Fidelity, Efficiency, Adaptivity
& Test ACC, Test loss; Time (s)
& MNIST, fMNIST, CIFAR10, CelebA & Non-IID 
& n/d & LeNet5, ResNet18 & - \\ \midrule

\cite{wang2024server} & Remaining clients & Sample
& Reverse Training: GA
& Post-training
& Efficacy, Fidelity
& Remaining Acc., loss; Time (s)
& MNIST, fMNIST, CIFAR10, CelebA & Non-IID
& FedAvg & MLP, LeNet5, ResNet18, MobileNet & -\\ \midrule

\cite{liu2024guaranteeing} & Remaining clients & Client
& Retrain the low-performing cluster
& -
& Efficacy, Fidelity, Efficiency, Guarantee, Privacy
& Remaining Acc.; Time (s), Comm (MB)
& MNIST, CIFAR10 & n/d
& SecAgg & 2-layer CNN, ResNet18 & - \\ \midrule

\cite{huynh2024fast} & Server & Client
& Subtract differences between target model updates and the aggregation of selectively stored local updates
& -
& Efficacy, Efficiency, Fidelity, Guarantee
& ASR (BA), Cosine Similarity; Remaining Acc.; Time (s), Mem (gb)
& MNIST, CIFAR10, OCTMNIST, MEDNMNISTv2 & Non-IID (Dirichlet)
& FedAvg & n/d & \href{https://github.com/thanhtrunghuynh93/fastFedUL}{link} \\ \midrule

\cite{su2024f2ul} & Server, Target client & Client
& Reverse training: GA
& Post-training
& Efficacy, Fidelity, Efficiency, Fairness
& Target Acc.; Remaining Acc., Test Acc.; Time (s, flops)
& MNIST, fMNIST, KMNIST, CIFAR10 & IID, Non-IID
& FedAvg & LeNet, AlexNet & - \\ \midrule

\cite{gu2024ferrari} & Server, Target Client & Features 
& Guided optimization to minimize feature sensitivity
& -
& Efficacy, Fidelity, Efficiency, Guarantee
& ASR (MIA), Target Acc., Attention map; Remaining Acc., Test Acc; Time (s, Flops)
& MNIST, fMNIST, CIFAR10, CIFAR100, CMNIST, CelebA, ImageNet, Adult, Diabetes, IMDb & IID, Non-IID (Dirichlet 1, 10)
& n/d & ResNet18, BERT & \href{https://github.com/OngWinKent/Federated-Feature-Unlearning}{link} \\ \midrule

\cite{xu2024update} & Remaining Clients & Class
& Finetuning the most influential channel identifying the target class 
& - 
& Efficacy, Fidelity, Efficiency 
& ASR (MIA), Target Acc., Grad Inversion Attack; Target Acc., Remaining Acc.; Time (s, round), Mem (MB)
& MNIST, fMNIST, CIFAR10, CIFAR100 & IID, Non-IId
& FedAvg & ResNet20, VGG11 & - \\ \midrule

\cite{deng2024enable} & Target Client & Class 
& Contrastive unlearning to make it similar to a model trained only on the retaining set at feature level
& Post-training on low-frequency components
& Efficacy, Fidelity, Efficiency
& Target Error; Remaining F1, Acc, Error, Test Acc, Error; Time (s)
& RSNA-ICH, ISIC2018 & Non-IID (Dirichlet 1)
& FedAvg & DenseNet121 & \href{https://github.com/dzp2095/FCU}{link} \\ \midrule

\cite{zhou2024streamlined} & Server & Sample
& KD: negative transfer knowledge to be forgotten 
& KD: transfer retaining knowledge
& Efficacy, Fidelity, Efficiency, Guarantee
& ASR (BA), Target Acc.; Remaining Acc.; Time (round)
& CIFAR10, CIFAR100, DBpedia & IID, Non-IID (Random)
& FedAvg & ResNst18, ResNet44, LSTM & - \\ \midrule

\cite{gu2024unlearning} & Server, Target Client & Sample, Class, Client
& Add auxiliary unlearning module while training and XOR with the local model upon unlearning request
& -
& Efficacy, Fidelity, Efficiency, Guarantee
& ASR (BA, MIA), Target Acc.; Remaining Acc.; Time
& MNIST, CIFAR10, CIFAR100 & IID, Non-IID (Dirichlet 1, 10)
& n/d & LeNet, AlexNet, ResNet18 & \href{https://github.com/Liar-Mask/FedAU}{link} \\ \midrule

\cite{pan2024federated} & Server & Client
& Reverse training: GA to orthogonal steepest descent direction
& Post-training with gradient projection
& Efficacy, Fidelity, Efficiency, Guarantee
& ASR (BA); Test Acc.; Time (round) 
& MNIST, fMNIST, CIFAR10, CIFAR100 & IID, Non-IID (Unique)
& FedAvg & LeNet5, MLP, 2-layer CNN, NFResnet18 & - \\ 

    \bottomrule
    \end{tabular}
\end{sidewaystable*}

\makeatletter \@rot@twosidefalse \makeatother
\begin{sidewaystable*}
    \caption{A comprehensive table summarizing the literature on FU. (continued)}
    \label{tab:comprehensive3}
    \footnotesize
    \setlength\tabcolsep{2.2pt} 
    \begin{tabular}{C{0.3cm}L{1.5cm}L{1cm} 
    L{4cm}L{3.9cm} 
    L{2.5cm}L{2.4cm} 
    L{2.1cm}L{1.5cm} 
    C{1.3cm}C{1.7cm}C{0.5cm}} 
    \toprule
    \multicolumn{1}{c}{\multirow{2}*{\textbf{Ref.}}} & \multicolumn{1}{c}{\multirow{2}*{\textbf{Unlearner}}} & \multicolumn{1}{c}{\multirow{2}*{\textbf{Target}}}
    & \multicolumn{1}{c}{\multirow{2}*{\textbf{Influence removal}}} & \multicolumn{1}{c}{\multirow{2}*{\textbf{Performance recovery}}} 
    & \multicolumn{1}{c}{\multirow{2}*{\textbf{Implication}}} & \multicolumn{1}{c}{\multirow{2}*{\textbf{Eval. Metric}}}
    & \multicolumn{1}{c}{\multirow{2}*{\textbf{Dataset}}} & \multicolumn{1}{c}{\multirow{2}*{\textbf{Data Dist.}}} 
    & \textbf{Agg. Method} & \textbf{Model Arch.} & \multicolumn{1}{c}{\multirow{2}*{\textbf{Code}}} 
    \\ \toprule 

\cite{wang2024fedu} & Target client & Class
& Approximate influence using Hessian-vector product and subtract it from a global model
& local post-training as an adaptive optimization
& Efficacy, Fidelity, Efficiency, Guarantee
& ASR (BA); Remaining Acc., Test ACC., Time (s)
& MNIST, CIFAR10, STL10 & IID
& n/d & 5-layer CNN, ResNet18 & - \\ \midrule

\cite{meerza2024confuse} & Target client & Sample, Client, Feature
& Create a fake dataset to confuse the model and penalize correct prediction by adding reg term
& Update salient weights only
& Efficacy, Fidelity, Fairness
& ASR (MIA, BA); Remaining Acc.
& MNIST, CIFAR10, Adult & n/d
& n/d & LeNet5, ResNet18, MLP & - \\ \midrule

\cite{li2024federated} & Server, Target client & Client
& Compute Approx. Hessian matrix to estimate recovered gradient 
& Gradient clipping
& Efficacy, Fidelity, Adaptivity
& ASR (BA); Test Acc.
& MNIST, GTRSB & n/d
& FedAvg & CNN & - \\ \midrule

\cite{zuo2024federated} & Remaining clients & Sample, Class, Client
& Rollback before the target join
& Retraining in other clients
& Efficacy, Efficiency, Security 
& Target ACC., loss; Test ACC, Remaining Acc; Time (s)
& MNIST, CIFAR10 & n/d 
& n/d & n/d & - \\ \midrule

\cite{ma2024hier} & Target client & Client
& Clustering, GA on target device 
& Post-training
& Efficacy, Fidelity, Efficiency, Guarantee, Scalability
& Target ACC; Test Acc, Remaining ACC, Time (s)
& fMNIST, CIFAR10, CIFAR100 & Non-IID (Random)
& FedAvg & MLP, AlexNet, ResNet18 & - \\ \midrule

\cite{yu2024federated} & Server, Remaining client & Client
& Oversampling from remaining clients to address bias, removing noise from the generated data
& Post-training on the oversampled data
& Efficacy, Fidelity, Fairness
& Target Acc; Test Acc
& MNIST, fMNIST, USPS & Covariate shift
& Weighted Avg & MLP, 4-layer CNN &-\\ \midrule

\cite{guo2024forgetting} & All clients & Class
& Multi-task learning; change the class of the target sample
& Post-training: fine-tuning with the new dataset
& Efficacy, Fidelity, Efficiency, Privacy
& Target Acc, F1, ASR (MIA); Remaining Acc, F1, Test Acc; Speedup ratio
& CIFAR10, CIFAR100, fMNIST, EuroSAT & IID, Non-IID (Dirichlet 0.1, 0.5)
& n/d & ResNet18, ResNet50 & -\\ \midrule

\cite{zhu2024federated} & Server & Client
& Client partitioning based on their distribution, remove the subset model
& -
& Efficacy, Fidelity, Efficiency, Scalability
& Activation mapping; test Acc, Time (s)
& MNIST, fMNIST, CIFAR10, SVHN & -
& Selective Avg & 3-layer CNN, AlexNet, ResNet50, VGG16 & - \\ \midrule

\cite{liu2024blockful} & Target client & Class, Client
& Reverse Learning (GA) 
& - 
& Efficacy, Fidelity, Efficiency
& Target Acc; Remaining Acc; time (s)
& fMNIST, CIFAR10 & n/d
& n/d & AlexNet, ResNet18, MobileNetv2 &- \\ \midrule

\cite{xie2024adaptive} & Server & Client
& Compute FIM to approximate the parameter and clip them
& KD: Using synthesized pseudo samples obtained from the feature space 
& Efficacy, Fidelity, Efficacy
& MIA Prec., Recall; Test Acc; Runtime (s)
& MNIST, CIFAR10, CIFAR100 & n/d
& n/d & LeNet5, ResNet18, ResNet34 & - \\ \midrule

\cite{mora2024fedunran} & Target Client & Client
& Replace the true label in the forget set with random ones
& Post-training with the modified dataset
& Efficacy, Fidelity, Efficiency
& ASR (MIA), Target Acc; Test Acc; Time (round)
& fMNIST & Semi Non-IID (Unique)
& FedAvg & LeNet & \href{https://github.com/alessiomora/FedUNRAN}{link}\\ \midrule

\cite{zuo2024federated} & Server & Client
& Unlearning using LoRA 
& -
& Efficiency, Efficacy
& Target Acc; time (s)
& IMDb, Twitter & Non-IID
& n/d & GPT-2 & - \\ \midrule

\cite{jiang2024feduhb} & Remaining clients & Client
& Rapid retraining: accelerating using heavy ball method
& - 
& Efficacy, Fidelity, Efficiency, Guarantee
& ASR (MIA, BA); Teset loss, Time (s)
& MNIST, CIFAR10 & n/d
& FedAvg & 3-layer CNN & -\\ \midrule

\cite{mora2024fedquit} & Target client & Client
& KD: transfer random prediction on forget data
& KD: transfer retaining knowledge
& Efficacy, Fidelity
& ASR (MIA), Target Acc; Remaining Acc, Test Acc, prediction overlap
& CIFAR10, CIFAR100, Non-IID (Dirichlet 0.1)
& n/d & ResNet18 & -\\ \midrule

\cite{xiong2024appro} & Target client & Client
& Compute approx hessian with remaining clients
& - 
& Efficacy, fidelity, Efficiency
& ASR (MIA), SAPE; Test loss, Model difference
& fMNIST, CIFAR10 & n/d
& n/d & 3-, 4-layer CNN & -\\ 

    \bottomrule
    \end{tabular}
\end{sidewaystable*}

\makeatletter \@rot@twosidefalse \makeatother
\begin{sidewaystable*}
    \caption{A comprehensive table summarizing the literature on FU. (continued)}
    \label{tab:comprehensive3}
    \footnotesize
    \setlength\tabcolsep{2.2pt} 
    \begin{tabular}{C{0.3cm}L{1.5cm}L{1cm} 
    L{4cm}L{3.9cm} 
    L{2.5cm}L{2.4cm} 
    L{2.1cm}L{1.5cm} 
    C{1.3cm}C{1.7cm}C{0.5cm}} 
    \toprule
    \multicolumn{1}{c}{\multirow{2}*{\textbf{Ref.}}} & \multicolumn{1}{c}{\multirow{2}*{\textbf{Unlearner}}} & \multicolumn{1}{c}{\multirow{2}*{\textbf{Target}}}
    & \multicolumn{1}{c}{\multirow{2}*{\textbf{Influence removal}}} & \multicolumn{1}{c}{\multirow{2}*{\textbf{Performance recovery}}} 
    & \multicolumn{1}{c}{\multirow{2}*{\textbf{Implication}}} & \multicolumn{1}{c}{\multirow{2}*{\textbf{Eval. Metric}}}
    & \multicolumn{1}{c}{\multirow{2}*{\textbf{Dataset}}} & \multicolumn{1}{c}{\multirow{2}*{\textbf{Data Dist.}}} 
    & \textbf{Agg. Method} & \textbf{Model Arch.} & \multicolumn{1}{c}{\multirow{2}*{\textbf{Code}}} 
    \\ \toprule 
    
\cite{khalil2025not} & Server & Class
& Subtract layer-wise parameters of the global model 
& Post-training
& Efficacy, Fidelity, Efficiency 
& Target Acc, ASR (MIA); Remaining Acc, Test Acc; Time (flops, bytes)
& Caltech101, CIFAR10, CIFAR100 & IID, Non-IID (Dirichlet 0.1)
& FedAvg & ViT-b/16, 2-layer CNN, ResNet18 & -\\ \midrule

\cite{wu2024unlearning} & Server & Client
& Recover from historically saved remaining client updates
& KD: transfer remembered knowledge
& Efficacy, Fidelity, Efficacy
& ASR (BA); Test Acc.; Time (s)
& MNIST, CIFAR10, CTSRB, IMDb & IID, Non-IID (Dirichlet [0.1, 1])
& FedAvg & 3-layer CNN, VGG9, 16, ResNet, MobileNetv2, AlexNet & - \\ \midrule

\cite{han2025vertical} & Target client & Client
& Reverse training: GA guided by Avged clean clients
& Post-training
& Efficacy, Fidelity, Efficiency
& Target Acc, ASR (BA), Recall (MIA); Remaining Acc,; time (s)
& MNIST, fMNIST, CIFAR10 & n/d
& n/d & CNN, AlexNet & \href{https://github.com/mengde-han/VFL-unlearn}{link} \\ \midrule

\cite{zhang2025model} & Server & Client
& Clip-guided few shot KD using a global model as a student
& distill only retained knowledge
& Efficacy, Fidelity, Efficiency
& Target Acc, Remaining Acc, loss, prediction probs on test set, Grad Cam
& MNIST, SVHN, CIFAR10, CIFAR100 & IDD, Non-IID (Dirichlet 0.1)
& FedAvg & LeNet5, ResNet18, ResNet50 & -\\ \midrule

\cite{wang2025poisoning} & Server & Client
& Compute approx. hessian matrix using limited historical data
& - 
& Efficacy, Fidelity, Security, Guarantee
& Poisoning ASR, Test Error
& MNIST & Non-IID (Dirichlet 0.5)
& FedAvg, Median, Trmean & 3-layer CNN & -\\ \midrule

\cite{varshney2025unlearningclientsfeaturessamples} & All clients & Sample, Client, Feature
& GA to unlearn sample, KD to unlearn features and clients
& - 
& Efficacy, Fidelity, Efficiency
& ASR (MIA); test loss, Remaining F1, AUC; Mem in GB
& UCI, CIFAR10, STL10 & n/d 
& n/d & MLP, ResNet18 & -\\ \midrule

\cite{Zhong2025UnlearningTK} & Server, Remaining clients & Sample, Class, Client
& Identify critical layer and mask it out
& -
& Efficacy, Fidelity, Efficiency, Guarantee
& Target Acc, ASR (MIA); Test Acc, Remaining Acc; Time (round), Mem (byte)
& fMNIST, CIFAR10, CIFAR100 & Non-IID (Dirichlet 1, 0.1)
& FedAvg & LeNet, ResNEt18, SimpleViT & -\\ \midrule

\cite{wang2025forgettingdatatimetheoretically} & Server & Sample, Feature
& Subtract target confidence from the original confidence matrix
& -
& Adaptivity, Privacy, Fidelity, Efficiency, Guarantee
& Test Acc, AUC; Time (epoch), Mem (Byte)
& Adult, Credit, Diabetes, Nursery, Malware & n/d
& n/d & n/d & \href{https://github.com/wangln19/vertical-federated-unlearning}{link} \\ \midrule

\cite{pan2025feature} & Server & Feature
& Dynamically add perturbation to target client updates 
& Gradient alignment
& Efficacy, Fidelity, Efficiency, Guarantee, Adaptibility
& Target ACC, ASR (MIA); Remaining F1, Acc; Time (epoch)
& Bank, Credit & IID, Non-IID (Random)
& Weighted Avg & CNN, Shallow CNN & -\\ \midrule

\cite{chen2025fedmua} & Target Clients & Client
& Identify influential samples, modify the prediction for the target samples
& -
& Efficacy, Fidelity
& Target ACC, ASR (MIA), Remaining ACC
& Purchase, MNIST, CIFAR10, CIFAR100, Credit & IID, Non-IID (Dirichlet 0.5)
& FedAvg & ResNet18, LeNet5, 4-layer CNN & -\\ \midrule

\cite{huynh2025certified} & Server & Client
& Rollback, Recover from sampled historical updates
& - 
& Efficacy, Fidelity, Guarantee
& ASR (Poisoning); HR@10, NDCG@10, ER@10, Cosine Sim., Time (s), Mem (gb)
& Movielens-1m, Pinterest & Covariate shift
& FedAvg & LightGCN, NCF & -\\

    \bottomrule
    \end{tabular}
\end{sidewaystable*}


}


 



\end{document}